\documentclass[journal=jacsat,manuscript=article]{achemso}

\usepackage[version=3]{mhchem} 
\usepackage{caption}
\usepackage{subcaption}
\usepackage{xcolor} 
\usepackage{mathtools}
\usepackage[toc]{appendix}
\usepackage{amssymb}
\usepackage{listings}

\usepackage{etoolbox} 

\usepackage{pdfpages}

\makeatletter
\patchcmd{\acs@contact@details}{E}{*\,E}{}{}
\patchcmd{\acs@email@list@aux}{;}{\par*\,Email}{}{} 
\makeatother

\setkeys{acs}{doi = true}

\SectionNumbersOn

\DeclareMathOperator*{\argmin}{argmin}
\let\bs\boldsymbol

\newtheorem{remark}{Remark}

\author{Madhav R. Muthyala}
\affiliation[osu]
{Department of Chemical and Biomolecular Engineering, The Ohio State University, Columbus, Ohio 43210, United States}

\author{Farshud Sorourifar}
\affiliation[osu]
{Department of Chemical and Biomolecular Engineering, The Ohio State University, Columbus, Ohio 43210, United States}

\author{You Peng}
\affiliation[dow]
{Chemometrics, AI and Statistics, Technical Expertise and Support, The Dow Chemical Company, Lake Jackson, Texas 77566, United States}

\author{Joel A. Paulson}
\email{paulson.82@osu.edu}
\affiliation[osu]
{Department of Chemical and Biomolecular Engineering, The Ohio State University, Columbus, Ohio 43210, United States}

\title[]{SyMANTIC: An Efficient Symbolic Regression Method for Interpretable and Parsimonious Model Discovery in Science and Beyond}


\usepackage{subfiles} 
\begin{document}

\begin{tocentry}
\includegraphics[width=\textwidth]{ 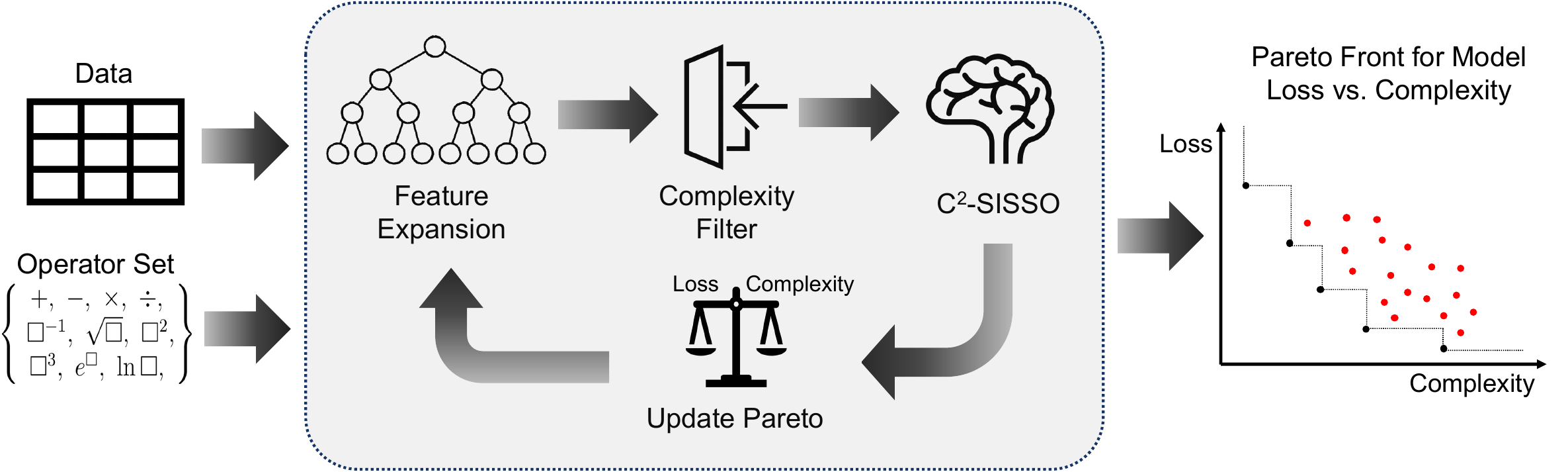}
\end{tocentry}

\clearpage

\begin{abstract}
Symbolic regression (SR) is an emerging branch of machine learning focused on discovering simple and interpretable mathematical expressions from data. Although a wide-variety of SR methods have been developed, they often face challenges such as high computational cost, poor scalability with respect to the number of input dimensions, fragility to noise, and an inability to balance accuracy and complexity. This work introduces SyMANTIC, a novel SR algorithm that addresses these challenges. SyMANTIC efficiently identifies (potentially several) low-dimensional descriptors from a large set of candidates (from $\sim 10^5$ to $\sim 10^{10}$ or more) through a unique combination of mutual information-based feature selection, adaptive feature expansion, and recursively applied $\ell_0$-based sparse regression. In addition, it employs an information-theoretic measure to produce an approximate set of Pareto-optimal equations, each offering the best-found accuracy for a given complexity. Furthermore, our open-source implementation of SyMANTIC, built on the PyTorch ecosystem, facilitates easy installation and GPU acceleration. We demonstrate the effectiveness of SyMANTIC across a range of problems, including synthetic examples, scientific benchmarks, real-world material property predictions, and chaotic dynamical system identification from small datasets. Extensive comparisons show that SyMANTIC uncovers similar or more accurate models at a fraction of the cost of existing SR methods.
\end{abstract}

\clearpage
\section{Introduction}
\label{sec:introduction}

Symbolic regression (SR) is a class of machine learning methods focused on the discovery of a symbolic expression (or model) that accurately fits a given dataset \cite{schmidt2009distilling, la2021contemporary, makke2024interpretable}. As an example, if one has measurements of planet masses $m_1$ and $m_2$, the distance between their centers of masses $r$, and the gravitational force acting between the two planets $F$, an SR algorithm would ideally ``re-discover'' Newton's law of universal gravitation $F = G \frac{m_1 m_2}{r^2}$ (where $G = 6.6743 \times 10^{-11}$ is the gravitational constant) by identifying the right combination of simple mathematical operations with the available variables. Learning models in the form of simple mathematical expressions offers much more than just potentially improved predictive power (via better generalization performance); it also enhances human interpretability and allows for deeper analysis, making these models ideal for use in high-stakes applications that have strict requirements on safety and reliability. Consequently, SR has been applied across many diverse fields including physics \cite{lemos2023rediscovering}, chemistry \cite{hernandez2019fast}, material science \cite{wang2019symbolic}, economics \cite{truscott2014explaining}, medicine \cite{virgolin2020machine}, control of dynamic systems \cite{derner2020constructing}, and space exploration \cite{martens2022symbolic}, to name a few. 

SR is a supervised learning task where the model space consists entirely of analytic expressions. Early SR efforts, particularly in scientific contexts, involved manually identifying simple yet accurate expressions through intuition and trial-and-error. A classic example is Johannes Kepler’s discovery in 1601 that Mars' orbit was an ellipse, a breakthrough achieved after four years of failed attempts and extensive analysis of planetary data \cite{koyre2013astronomical}. Modern SR algorithms aim to automate this discovery process by harnessing the power of modern computers, allowing for the exploration of a far larger set of expressions than what could be manually evaluated by humans. However, automatically finding the exact solution to general SR problems is extremely difficult and time-consuming. In fact, SR has recently been shown to be an NP-hard problem \cite{virgolin2022symbolic}, even in the context of low-dimensional datasets. This difficulty arises from the fact that the space of analytic expressions grows exponentially large with increasing input size; note that a similar combinatorial challenge is the root of difficulty in solving many famous problems such as codebreaking, the Rubik's cube, and identifying the genetic codes that produce the most fit organisms.

\subsection{Context and Scope of Symbolic Regression}

The focus of this paper is on SR for algebraic models of the form $y = f(\bs{x}) + \epsilon$, where $y$ is the target variable, $\bs{x}$ are input features, and $\epsilon$ represents noise. This type of SR aligns closely with standard supervised learning tasks and is the primary scope of our work.

SR has also been extended to dynamic systems, where the goal is to uncover differential equations governing the evolution of system states over time, e.g., $\dot{\bs{z}}(t) = \bs{f}(\bs{z}(t))$. Methods such as SINDy \cite{brunton2016discovering} handle this by assuming access to both state measurements $\bs{z}(t)$ and their derivatives $\dot{\bs{z}}(t)$. Under this assumption, virtually any SR method can then be used to learn a representation of the dynamics $\bs{f}$. When derivatives are not available, one must resort to a more advanced approach beyond standard SR such as numerical differentiation \cite{chartrand2011numerical}, integration-based error minimization \cite{chen2018neural}, or dynamic optimization \cite{lejarza2022data}. While we briefly touch on the application of our approach to dynamic systems in Section \ref{subsubsec:dynamics}, our focus remains on algebraic models such that a comprehensive review of more recent literature related to learning differential equations from data is outside the scope of this work.

\subsection{Existing Approaches to Symbolic Regression}

We broadly categorize symbolic regression (SR) methods into four classes: brute-force search, genetic programming, deep learning-based methods, and sparse regression techniques. Each of these approaches has distinct advantages, such as flexibility, scalability, or interpretability, but also faces limitations that impact their applicability to different problem settings. We discuss these strengths and limitations in more detail in the paragraphs below.

\paragraph{Brute-force search} Brute-force search methods systematically enumerate possible mathematical expressions to identify the best fit for data. Early systems like BACON \cite{langley1977bacon} and FAHRENHEIT \cite{langley1989data} successfully discovered simple physical laws from idealized datasets but relied heavily on heuristics, limiting their applicability to more complex or nonlinear systems. These approaches laid foundational groundwork but are no longer practical due to the exponential growth of search spaces with increasing complexity.

\paragraph{Genetic programming} Genetic programming (GP) \cite{koza1994genetic} is one of the most widely used SR techniques. It evolves populations of symbolic expressions using biologically inspired operations such as mutation, crossover, and selection. GP's flexibility allows it to explore vast model spaces under minimal prior assumptions, making it powerful for discovering nonlinear relationships. However, GP often suffers from ``bloat,'' where successive generations produce overly complex models, hampering interpretability and increasing computational cost. Modern frameworks such as PySR \cite{cranmer2023interpretable}, GP-GOMEA \cite{virgolin2021improving}, and QLattice \cite{brolos2021approach} introduce optimizations to improve efficiency and scalability, including symbolic simplifications, multi-objective optimization, and Bayesian-inspired search strategies. Despite these advancements, GP methods remain computationally intensive and require extensive hyperparameter tuning, particularly for high-dimensional datasets.

\paragraph{Deep learning-based approaches} Deep learning has inspired SR methods that leverage neural networks to discover symbolic relationships. Methods such as Deep Symbolic Optimization (DSO) \cite{petersen2021deep} and Equation Learner (EQL) \cite{sahoo2018learning} use neural networks to represent and optimize symbolic expressions, while SR-Transformer \cite{kamienny2022endtoendsymbolicregressiontransformers} adopts a sequence-to-sequence architecture inspired by transformers to generate symbolic models. Hybrid approaches like AI-Feynman \cite{udrescu2020ai} combine neural network training with brute-force search, leveraging domain knowledge to decompose and simplify problems. While deep learning methods are effective for modeling complex systems, they often rely on large datasets, suffer from interpretability challenges, and struggle in the presence of noisy or sparse data.

\paragraph{Sparse regression techniques} Sparse regression methods tackle SR problems by exploiting the assumption that only a small number of terms are required to describe the underlying data relationships. One class of methods formulates SR as a best subset selection problem, where the goal is to identify the optimal combination of features from a predefined library to construct the most parsimonious model. Mixed-integer programming (MIP) techniques solve this problem by explicitly enforcing sparsity constraints through binary variables. For example, ALAMO \cite{cozad2014learning} uses a mixed-integer linear programming (MILP) approach to identify the best subset of terms, limiting the model to a user-set maximum number of non-zero coefficients. More recent methods \cite{austel2017globally, cozad2018global} extend this formulation to mixed-integer nonlinear programming (MINLP), allowing for the discovery of more complex symbolic expressions. However, the computational cost of MINLP formulations grows exponentially with the size of the feature space and the number of data points, making these methods impractical for high-dimensional problems unless significant feature pruning is performed beforehand. This greatly limits the space of functions that can be effectively explored by these methods.

Another class of sparse regression methods takes an approximate approach to enforcing sparsity, using regularization penalties to encourage sparse solutions. These include techniques such as LASSO ($\ell_1$ regularization) and elastic net, which aim to reduce the number of active features while maintaining computational efficiency. LASSO penalizes the absolute value of coefficients to shrink many toward zero, while elastic net combines $\ell_1$ and $\ell_2$ penalties to handle correlated features better. While these methods are computationally efficient, they require careful tuning of regularization parameters, which can be nontrivial, and may include too many features, limiting the simplicity/interpretability of the resulting models.

To address these limitations, the Sure Independence Screening and Sparsifying Operator (SISSO) \cite{ouyang2018sisso} method combines feature selection and sparse regression in a two-step process. SISSO first uses Sure Independence Screening (SIS) to efficiently reduce the feature space by selecting terms with strong marginal correlations to the target variable. It then applies $\ell_0$ regularization within the reduced feature space to identify the optimal subset of terms. This approach balances computational cost and accuracy, allowing SISSO to scale to high-dimensional problems while maintaining interpretability. However, SISSO requires users to manually tune hyperparameters, such as the number of retained features and sparsity levels. While recent advancements like the \texttt{TorchSISSO} package \cite{muthyala2024torchsisso} have introduced GPU acceleration, challenges remain, including the lack of automation and the inability to return multiple plausible models that tradeoff accuracy and complexity.
\subsection{Contributions of This Work}

As highlighted in the previous section, existing SR methods face challenges such as high computational cost, difficulty handling high-dimensional inputs, sensitivity to noise, limited exploration of complexity-accuracy tradeoffs, and a lack of user-friendly, open-source implementations. To address these gaps, we propose SyMANTIC (Symbolic Modeling with Adaptive iNtelligent feaTure expansIon), a new SR algorithm and Python package that builds on the foundation of SISSO \cite{ouyang2018sisso} and introduces several novel features. The main contributions of this work are as follows:

\begin{itemize} 
\item \textbf{Automated feature screening:} SyMANTIC incorporates a feature screening step that efficiently prunes the primary feature space before recursive feature expansion, improving its ability to scale to datasets with a large number of inputs.
\item \textbf{Information-theoretic complexity measure:} SyMANTIC introduces an information-theoretic complexity metric inspired by the minimum description length principle \cite{hansen2001model}. This approach improves robustness compared to simpler metrics, such as total operation count, by accounting for both the number and types of operations performed. 
\item \textbf{Pareto frontier construction:} By explicitly tracking an approximate Pareto frontier of models, SyMANTIC provides users with a systematic way to evaluate tradeoffs between model complexity and predictive performance. 
\item \textbf{Automated hyperparameter tuning:} To reduce the burden on users, SyMANTIC automates the tuning of key hyperparameters, such as sparsity levels and screening thresholds, improving usability and reducing reliance on trial-and-error. 
\item \textbf{GPU-accelerated implementation:} Built on PyTorch \cite{paszke2019pytorch}, SyMANTIC supports batched computations and GPU acceleration, significantly reducing computation times for more complicated models and larger datasets. 
\item \textbf{Comprehensive evaluation:} SyMANTIC outperforms state-of-the-art SR methods across diverse benchmark problems, demonstrating superior accuracy, scalability, and robustness to noise on synthetic and real-world problems.
\end{itemize}

The Python implementation of SyMANTIC is available as an open-source package on PyPI under the name \texttt{symantic}, making it widely accessible to the research community.

\subsection{Organization of the Paper}

The remainder of this paper is organized as follows. In Section \ref{sec:symantic-method}, we provide a mathematical description of the SR problem and our proposed SyMANTIC algorithm. We also provide a brief overview of our \texttt{symantic} Python package and some important practical considerations. In Section \ref{sec:case-studies}, we compare SyMANTIC to several popular SR packages on a variety of different problems ranging from synthetic and science benchmarks to nonlinear system identification and material property modeling. We observe SyMANTIC can outperform all tested competing methods (including those recently reported to be at or near the state of the art based on extensive benchmarking \cite{la2021contemporary}) in both accuracy and the time required to find a solution. Lastly, we provide some concluding remarks in Section \ref{sec:conclusion}. 

\section{The SyMANTIC Method}
\label{sec:symantic-method}

\subsection{Problem Description}
\label{subsec:prob-description}

We can formulate the SR problem as a type of empirical risk minimization problem \cite{makke2024interpretable} over some class of functions $\mathcal{F}$ given a training dataset $\mathcal{D}$
\begin{align} \label{eq:sr-problem}
    f^\star = \argmin_{f \in \mathcal{F}} ~ L(f) = \frac{1}{n}\sum_{i=1}^n \ell( f(\bs{x}_i), y_i ),
\end{align}
where $\mathcal{D} = \{ (\bs{x}_i, y_i) \}_{i=1}^n$ is composed of input $\bs{x}_i = (x_{1,i}, \ldots, x_{D,i}) \in \mathbb{R}^D$ and scalar output $y_i \in \mathbb{R}$ pairs, $\mathcal{F}$ consists of valid functions/mappings $f : \mathbb{R}^D \to \mathbb{R}$, $l(\cdot)$ is the loss function, $L(\cdot)$ is the expected loss (also known as empirical risk) measure, and $f^\star$ is the optimal model that produces the lowest average loss across the training data. The loss function is typically chosen to be the negative log-likelihood (NLL) such that \eqref{eq:sr-problem} can be interpreted within the maximum likelihood estimation (MLE) framework. A common choice for the loss is the mean squared error (MSE), i.e., $\ell( f(\bs{x}_i), y_i ) = (y_i - f(\bs{x}_i))^2$, which is the exact NLL when the measurement errors are additive and i.i.d. Gaussian random variables \cite{bengio2017deep}. In this work, we will assume an MSE loss function for simplicity, though the components of our proposed algorithm can be extended to handle more general choices of loss functions. 

A critical difference in SR compared to other regression methods is the choice of $\mathcal{F}$, which should be ``interpretable'' functions specified as a library obtained by function composition over elementary arithmetic operations and mathematical functions and variables. Specifically, $\mathcal{F}$ is defined as all functions that can be formed by composition of the elements of the \textit{primitive set} $\mathcal{P}$, which can contain a mix of functions that perform simple algebraic operations such as addition and subtraction; transcendental functions such as sine, cosine, and exponential; constant functions such as $c_{a}(\bs{x}) = a$ for some $a \in \mathbb{R}$; and identity functions that represent the variables of interest such as $I_1(\bs{x}) = x_1$ and $I_2(\bs{x}) = x_2$. For example, selecting $\mathcal{P} = \{ +(\cdot,\cdot), -(\cdot,\cdot), \times(\cdot,\cdot), x_1, x_2, +1, -1 \}$ implies $\mathcal{F}$ will contain the set of all two-dimensional polynomials of arbitrary degree in $x_1$ and $x_2$ with integer coefficients. Even though one can place further restrictions on $\mathcal{F}$, we can see that the size of this search space grows exponentially fast with the size of the primitive set for a fixed recursion depth. 
It is thus (effectively) impossible to exactly solve \eqref{eq:sr-problem} in all but the simplest cases since we must exhaustively test all possible functions in $\mathcal{F}$ via brute force \cite{virgolin2022symbolic}. 

Another complication is that, as is typically the case in learning problems, our actual goal is to discover a function $f$ that generalizes to new (unseen) observations that come from the same underlying distribution that generated the training set $\mathcal{D}$. Therefore, $f^\star$ should not only achieve minimal loss on $\mathcal{D}$ but also remain minimal for new observations that are not yet available (often referred to as a test dataset $\mathcal{D}_{\text{test}}$). A common way to deal with this challenge in practice, in the spirit of Occam’s razor, is to aim to find functions $f$ that jointly minimize the expected loss measure $L(f)$ and some measure of function complexity $C(f)$. This requires identifying \textit{Pareto-optimal} functions defined such that no other function in $\mathcal{F}$ exist that is both less complex and more accurate. Generating the exact Pareto frontier would require solving a family of regularized SR problems of the form $\min_{f \in \mathcal{F}} L(f) + \lambda C(f)$ for $\lambda \geq 0$. Since even a single problem with fixed $\lambda$ cannot be exactly solved, we must rely on certain assumptions/heuristics to develop a tractable algorithm that produces a reasonable approximation of the Pareto frontier. We discuss our proposed methodology next, which takes a sparse regression (or compressed sensing) view of penalized SR, with special attention to how function complexity and the function class are incorporated into the search.  

\subsection{Overall Algorithm}
\label{subsec:algorithm}

\begin{figure}[tb!]
    \centering
    \includegraphics[width=0.99\textwidth]{ 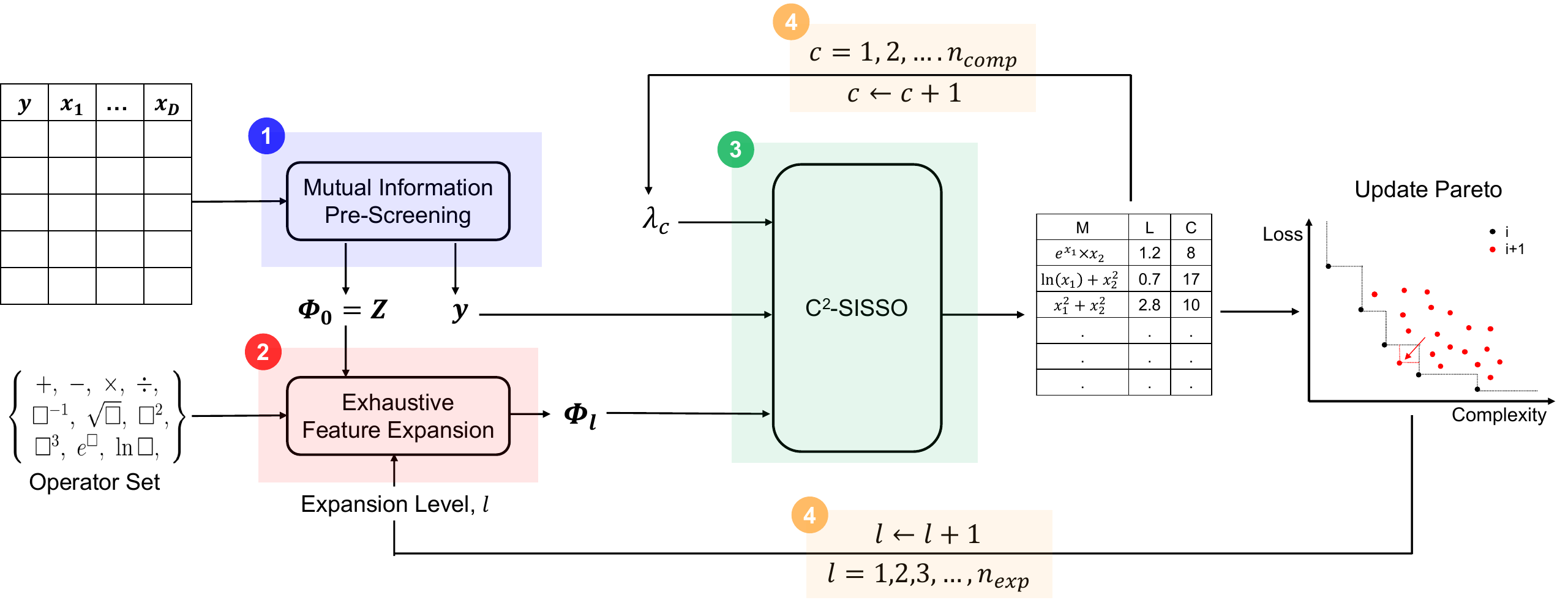}
    \caption{Schematic illustration of our proposed SyMANTIC algorithm. It takes in training data in the form of several $(\bs{x}_i, y_i)$ pairs and a user-defined operator set, with the default choice shown in \eqref{eq:operator-set-default}. The training data is fed to a mutual information pre-screening method (Step 1) to limit the number of primary features sent to feature expansion (Step 2). The feature expansion step is carried out recursively over some number of levels $l$. The expanded features, along with a complexity filter parameter $\lambda_c$, are input to our C$^2$-SISSO algorithm (Step 3) whose details are summarized in Figure \ref{fig:c2sisso-flowchart}. C$^2$-SISSO outputs a new set of tested models $\mathcal{M}$ with corresponding loss $\mathcal{L}$ and complexity $\mathcal{C}$ values. Using the new $\mathcal{L}$ and $\mathcal{C}$, the current approximation to the Pareto front is updated. We run an automated procedure over several $l$ and $c$ values (Step 4) to recursively improve the Pareto front. Note that the algorithm will stop and return the current Pareto front when either a threshold on the root mean squared error is satisfied or all hyperparameter combinations are tried (Exit Condition).}
    \label{fig:symantic-flowchart}
\end{figure}

The overall \textbf{SyMANTIC} (\textbf{Sy}mbolic \textbf{M}odeling with \textbf{A}daptive i\textbf{N}elligent fea\textbf{T}ure expans\textbf{I}on and \textbf{C}ompressed sensing) algorithm is schematically illustrated in Figure \ref{fig:symantic-flowchart}. Its core component is a module we refer to as C$^2$-SISSO (Complexity-Constrained Sure Independence Screening with Sparsifying Operator) that builds upon the SISSO framework, which is an efficient compressed sensing approach for identifying optimal low-dimensional descriptors from a massive set of candidates \cite{ouyang2018sisso}. SISSO was originally developed in the context of material property modeling, though the principles can be applied more generally to SR problems, which is the view we take in this work. Since we also care about the complexity of the generated equations, C$^2$-SISSO limits the feature complexity before running any sparse regression algorithm.
C$^2$-SISSO depends on a number of key hyperparameters such as the generated feature space and a complexity constraint bound. Not only can these quantities be difficult to set manually, the best models at different complexities are unlikely to be generated from the same hyperparameters. As such, SyMANTIC runs an automated procedure to sequentially test several combinations of hyperparameters -- after each run, the current estimate of the Pareto front between loss and complexity is updated with the new results from C$^2$-SISSO. 
Since the feature space sent to C$^2$-SISSO is recursively built from $\bs{x}$ and a user-defined operator set, its exponential growth with number of (binary) operators and number of (starting) primary features can quickly become prohibitive both in terms of computational cost and memory. Thus, SyMANTIC also incorporates a mutual information-based screening method to handle the high-dimensional case ($D$ larger than roughly 10 to 20), which is fairly robust to situations where the relationship between $\bs{x}$ and $y$ is strongly nonlinear. Below, we describe each of the main algorithmic steps of SyMANTIC in more detail. 

\paragraph{Mutual information-based feature pre-screening} SyMANTIC performs an initial variable screening method to the primary features $\bs{x}$ to limit the number that is sent to the adaptive feature expansion step. Since the relationship between $\bs{x}$ and $y$ can be highly nonlinear, we rely on a mutual information (MI) metric to measure the strength of the relationship between a given component $x_i$ and the target $y$ \cite{battiti1994using}
\begin{align} \label{eq:mi-exact}
    \text{MI}(y ; x_i) = \int p(x_i, y) \log\left( \frac{p(x_i, y)}{p(x_i) p(y)} \right)\text{d}x\text{d}y, 
\end{align}
where $p(x_i, y)$ is the joint probability density function between random variables $x_i$ and $y$ and $p(x_i)$ and $p(y)$ are their corresponding marginal distributions. The metric $\text{MI}(y ; x_i)$ is always non-negative and is equal to 0 if and only if $y$ and $x_i$ are independent. 
We rank the features according to their MI values by computing $m_i = \text{MI}(y ; x_i)$ for all $i = 1,\ldots, D$ and then applying the Argsort operator to these values, which returns the indices sorting the vector $\bs{m} = (m_1,\ldots, m_D)$ in descending order. Let $\sigma(\bs{m}) = (\sigma_1(\bs{m}), \ldots, \sigma_D(\bs{m})) = \text{Argsort}( \bs{m} )$ such that $m_{\sigma_1(\bs{m})} \geq m_{\sigma_2(\bs{m})} \geq \cdots \geq m_{\sigma_D(\bs{m})}$. Note that, if some of the coordinates of $\bs{m}$ are equal, we break ties arbitrarily.
The set of primary features sorted by MI is then denoted as $\bs{x}_{ \sigma(\bs{m}) }$. We apply two filters to select the final set of screened features that are passed to the rest of the SyMANTIC algorithm. Specifically, we drop all features that have a MI value below a certain threshold $\gamma$ and limit the number of screened features to be at most $n_\text{screen}$. Although $\gamma$ and $n_\text{screen}$ can be modified by users, we select default values of $\gamma = 0.1$ and $n_\text{screen} = 20$, as they were found to work well in our numerical experiments.
The final set of screened features $\bs{z} \in \mathbb{R}^d$ can then be expressed as follows 
\begin{align} \label{eq:mi-screen}
    \bs{z} = \bs{x}_{\mathcal{I}} ~~~ \text{where} ~~~ \mathcal{I} = \{ i \in \{1,\ldots,D\} : m_{\sigma_i(\bs{m})} \geq \gamma ~ \text{and} ~ i \leq n_\text{screen} \}.
\end{align}
In practice, we do not have direct access to the probability density functions  in \eqref{eq:mi-exact} such that, as commonly done, we use kernel density estimation \cite{chen2017tutorial} to compute estimates of MI.

\begin{remark}
The feature pre-screening step described above does not explicitly account for relationships or redundancy between primary features due to the computational cost of more sophisticated methods such as those based on conditional mutual information \cite{beraha2019feature}. To mitigate potential redundancy issues, we recommend that users perform an initial analysis of feature correlations to reduce the number of highly redundant features before applying \eqref{eq:mi-screen}. A simple and effective strategy is to compute the correlation matrix for the features and eliminate all but one feature from pairs with correlation coefficients above a chosen threshold (typically 0.9 or higher). When it is not obvious which feature to retain, users can explore running SyMANTIC on different subsets of features. Given the computational efficiency of SyMANTIC, such experiments are feasible and can help identify the most informative feature set.
\end{remark}

\paragraph{An information-theoretic measure of structural complexity} Although a variety of different metrics have been proposed to measure the complexity of an expression, we focus on one related to the \textit{minimum description length} \cite{hansen2001model}, which can be interpreted as a mathematical implementation of Occam's razor. Using the information-theoretic procedure developed by Wu and Tegmark \cite{wu2019toward}, one can define the complexity (in bits) of a function $f(\bs{x} ; \bs{\theta})$ with input $\bs{x}$ and parameters $\bs{\theta}$ as $K(f) \log_2 B(f) + \sum_{i} L_d(\theta_i)$ where $B(f)$ is the number of basis functions appearing in $f$, $K(f)$ is the number of times the basis function are used in the expression, and $L_d(c)$ is the description length of a constant (e.g., equal to $\log_2 (1 + |c|)$ when $c$ is an integer). It turns out that this measure of complexity is sensitive to the representation of the parameters. For example, from a user perspective, the expressions $0.500137 \times m \times v^2$ and $1/2 \times m \times v^2$ are virtually identical since the user can quickly identify that the real number $0.500137$ can be replaced with the rational number $1/2$, which requires a smaller number of bits to represent. Therefore, we propose to focus on the \textit{structural complexity} of the function, which amounts to only using the first term in the description length expression
\begin{align} \label{eq:structural-complexity}
    C(f) = K(f) \log_2 B(f).
\end{align}
Note that both input variables and mathematical functions count toward $B(f)$ and $K(f)$. For example, consider the expression $f(\bs{x}) = 0.5 \times x_1 \times x_2^2$. This expression contains $B(f) = 4$ basis functions ($x_1$, $x_2$, $\times(\cdot,\cdot)$, $\cdot^2$) that are used a total of $K(f) = 5$ times such that $C(f(\bs{x}) = 0.5 \times x_1 \times x_2^2) = 5 \log_2(4) = 10$. 

\paragraph{Construction of candidate function library} Given the set of screened features $\bs{z} \in \mathbb{R}^d$, we generate a large candidate set of nonlinear features by recursively applying functional and/or algebraic operations (referred to as operators) to $\bs{z}$. Let $\mathcal{O}$ denote the operator set, consisting of unary $o[z_i]$ and binary $o[z_i, z_j]$ operators that can be applied to elements of $\bs{z}$. The choice of $\mathcal{O}$ is completely open to the user and can include any computable basis functions, such as dilogarithm and Bessel functions that are known to appear in certain types of particle physics and transport phenomena problems, as well as custom binary operators. Our default choice for $\mathcal{O}$ is the following
\begin{align} \label{eq:operator-set-default}
    \mathcal{O}[z_i, z_j] = \{ I(z_i), z_i + z_j, z_i - z_j, z_i \times z_j, z_i / z_j, \exp(z_i), \log(z_i), \sqrt{z_i}, z_i^{-1}, z_i^2 \},
\end{align}
where $I$, $\exp$, and $\log$ are the standard identity, exponential, and logarithm operators. Let $\bs{\phi}_l(\bs{x}) \in \mathbb{R}^{d_l}$ denote the vector of features at the $l^\text{th}$ level of expansion with total number of elements equal to $d_l$, which is a function of the original feature vector $\bs{x}$. We define the feature vectors recursively by exhaustively applying the operator to all valid combinations of the features at the previous level $\bs{\phi}_{l-1}(\bs{x})$, i.e., 
\begin{align}
    \bs{\phi}_l(\bs{x}) = \{ \mathcal{O}[z_i(\bs{x}), z_j(\bs{x})], ~ \forall z_i(\bs{x}), z_j(\bs{x}) \in \bs{\phi}_{l-1}(\bs{x}) \} ~~~ \text{with} ~~~ \bs{\phi}_0(\bs{x}) = \bs{z}(\bs{x}). 
\end{align}
For $m_u$ unary operators and $m_b$ non-symmetric binary operators, the number of features at level $l > 0$ can be expressed as
\begin{align} \label{eq:growth-in-feature}
    d_l = m_u d_{l-1} + m_b d_{l-1}(d_{l-1} - 1),~~~ d_0 = d.
\end{align}
Since the quadratic term quickly dominates as $l$ increases, we see that $d_l$ roughly scales as $d_l \sim m_b d_{l-1}^2$ for $l > 0$. Expressing this scaling law in terms of the number screened primary features $d$, we see that $d_l \sim m_b^{2^l - 1} d^{2^l}$, meaning the size of $\bs{\phi}_l(\bs{x})$ grows exponentially with the number of levels $l$. Note that, without the MI screening approach, $d$ can easily become too large to expand past a single level. 

For a given level $l$, we aim to represent the set of possible functions as linear combinations of the nonlinearly expanded features $\bs{\phi}_l^\top(\bs{x}) \bs{c}_l$ where $\bs{c}_l \in \mathbb{R}^{d_l}$ is a vector of coefficients. We discuss how SyMANTIC searches over $l$ later in this section. Also, note that we assume a constant feature (of all ones) is included in $\bs{\phi}_l(\bs{x})$ that serves as the bias term in the model.
Although we could attempt to apply standard linear regression to estimate $\bs{c}_l$ from the available training data, whenever $d_l \geq n$, we have an underdetermined problem such that we cannot find a unique coefficient vector that minimizes the training loss. Furthermore, we do not expect all of these features to be needed to predict the target $y$ -- this belief comes from the fact that many physical problems have a sparse (low complexity) solution. Therefore, we must resort to sparse regression methods that incorporate a regularization term to limit the number of non-zero coefficients that appear in the estimated model. 

\paragraph{C$^2$-SISSO algorithm} SISSO \cite{ouyang2018sisso} is a framework for identifying a small set of descriptors out a large set of candidates. It is focused on obtaining low expected loss (high accuracy) models by exploring a limited combinations of ``active features'' (those with non-zero coefficients) from the pool $\bs{\phi}_l(\bs{x})$ at a given expansion level $l$. Complexity is thus measured by the zero norm $\| \bs{c}_l \|_0$ that is directly equal to the number of non-zero coefficients, which does not account for the complexity of the features themselves. We propose a simple modification of SISSO that incorporates an initial complexity constraint, called C$^2$-SISSO. 
We will describe a generic form of this algorithm as a function $(\mathcal{M}, \mathcal{L}, \mathcal{C}) = \text{C}^2\text{-SISSO}( \bs{y}, \bs{\Phi}, \lambda )$ that takes as input the training outputs $\bs{y} = (y_1, \ldots, y_n) \in \mathbb{R}^n$, a feature matrix $\bs{\Phi} \in \mathbb{R}^{n \times d}$ (corresponding to all $d$ features evaluated at the training inputs), and a parameter $\lambda \geq 0$ and returns the set of all tested models $\mathcal{M}$ with corresponding loss $\mathcal{L}$ and complexity $\mathcal{C}$ values. 

We first get the set of reduced-complexity features by removing all features with complexity larger than $\lambda$. Let $\tilde{\bs{\Phi}} = \bs{\Phi}_\mathcal{I}$ denote the resulting submatrix of $\bs{\Phi}$ obtained by extracting its columns corresponding to indices $\mathcal{I} = \{ 1 \leq i \leq d : C(\phi_i(\bs{x})) \leq \lambda \}$. We then apply sure independence screening (SIS) \cite{fan2008sure} to identify a further subset of features that are most correlated with the output. SIS identifies the top $k$ features by running a simple calculation on a given output vector $\bs{y}$ and feature matrix $\bs{\Phi}$ (assuming its columns have been standardized to have zero mean and unit variance). It first computes a $d$-dimensional vector of weights $\boldsymbol{\omega} = ( \omega_1, \ldots, \omega_d ) = \boldsymbol{\Phi}^\top \bs{y}$ and then identifies the indices with the top $k$ magnitude weight, i.e., $\mathcal{S} = \{ 1 \leq i \leq d : | \omega_i | ~ \text{is among the first $k$ largest of all} \}$. We use the following shorthand to denote the general SIS procedure $\mathcal{S} = \text{SIS}( \bs{y}, \bs{\Phi} )$.

\begin{figure}[tb!]
    \centering
    \includegraphics[width=0.99\textwidth]{ 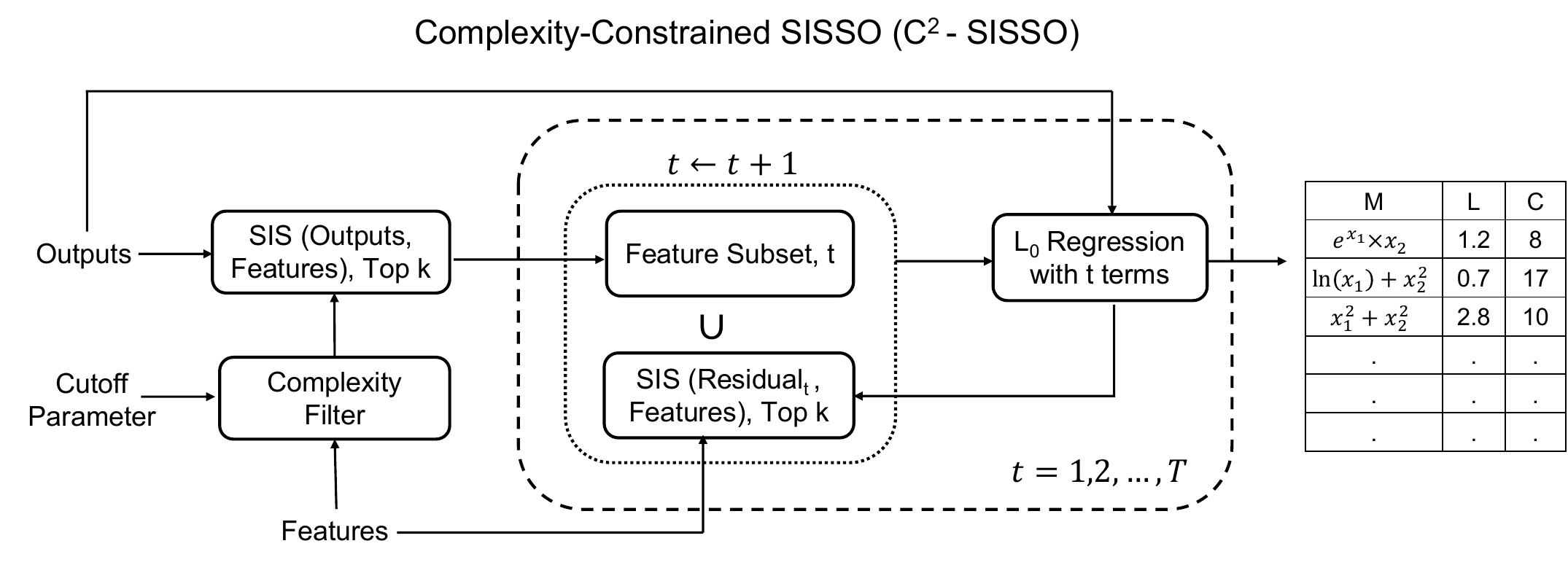}
    \caption{Schematic illustration of C$^2$-SISSO that takes as inputs features, measured outputs, and a complexity cutoff parameter and returns symbolic expression, loss, and complexity values for a set of tested models. The models are trained using $\ell_0$ regression to identify the best $t$ term model over a subset of features sequentially identified using sure independence screening (SIS) applied to the residual of the previous model. The process is repeated until a maximum number of terms $T$, which is set by the user.}
    \label{fig:c2sisso-flowchart}
\end{figure}

Given a subspace of the reduced-complexity feature matrix $\tilde{\bs{\Phi}}_\mathcal{S} \in \mathbb{R}^{n \times k}$ (with $k \ll d$), we can fit models using standard linear regression. However, it is still not clear how many non-zero coefficients we want to retain in the model. Thus, we sequentially build models from one to a maximum number of terms $T$, each time using the residual to guide an enlargement of the subspace. Let $\bs{r}_t \in \mathbb{R}^n$ denote the residual error for a $t$-term linear model (i.e., a model with $t$ non-zero coefficients) selected from a feature subspace $\mathcal{S}_t$. We can compute a closed-form expression for this error as $\bs{r}_t = \bs{y} - \tilde{\bs{\Phi}}_{\mathcal{S}_t} \bs{E}_t \bs{c}_t$ where $\bs{E}_t \in \mathbb{R}^{k \times t}$ is a binary matrix that selects $t$ feature columns out of the available feature space and $\bs{c}_t = ( \bs{E}_t^\top \tilde{\bs{\Phi}}_{\mathcal{S}_t}^\top \tilde{\bs{\Phi}}_{\mathcal{S}_t} \bs{E}_t )^\top \bs{E}_t^\top \tilde{\bs{\Phi}}_{\mathcal{S}_t}^\top \bs{y}$ is the coefficient vector obtained as the least squares solution from fitting $\tilde{\bs{\Phi}}_{\mathcal{S}_t} \bs{E}_t$ to $\bs{y}$. Note that this is an advantage to the MSE loss function, as it admits simple closed-form solutions that will be exploited in our software implementation. The recursive construction of the feature space can be mathematically written as 
\begin{align}
    \mathcal{S}_{t+1} = \mathcal{S}_{t} \cup \text{SIS}( \bs{r}^\star_{t}, \tilde{\bs{\Phi}} ), ~~ \forall t = 1, \ldots, T-1 ~~ \text{with} ~~ \mathcal{S}_{1} = \text{SIS}( \bs{y}, \tilde{\bs{\Phi}} ),
\end{align}
where $\bs{r}^\star_{t}$ is the residual error of the best model tested with $t$ terms from the subspace $\mathcal{S}_t$. To find $\bs{r}^\star_{t}$, we apply $\ell_0$ regression that can be carried out exactly for relatively small $T$. Specifically, we train all possible $t$ term models out of the reduced feature space (which corresponds to $tk \choose t$ models) and find the one with the lowest residual error. A schematic illustration of C$^2$-SISSO can be found in Figure \ref{fig:c2sisso-flowchart}.

One way to think about how SyMANTIC uses C$^2$-SISSO is as an efficient model exploration approach. Although the original SISSO work is focused on finding a model with loss lower than a threshold value, we are interested in testing a wider range of models to see where they land in the loss versus complexity landscape (more specifically near the estimated Pareto frontier). By design, a single run of our implementation of C$^2$-SISSO tests a total of $\sum_{t=1}^T {tk \choose t}$ models. Therefore, as opposed to just taking the best performing model out of these candidates, we return the loss and complexity of all of these models that can serve as an approximate Pareto frontier. The main challenge is that it is not always obvious how to select the right inputs to C$^2$-SISSO to produce a high-quality approximate Pareto front. Thus, SyMANTIC incorporates a simple but fully automated search procedure for these hyperparameters, described next. 

\begin{remark}
    We note that SIS \cite{fan2008sure} is conceptually similar to traditional feature selection techniques like forward and backward selection, which iteratively add or remove features based on how strongly they contribute to model performance. The key distinction, however, is computational scalability, as SIS only requires a simple matrix-vector multiplication to derive an importance score whereas forward/backward feature selection require a separate regression model to be trained per feature. This efficiency enables SIS to scale to huge feature spaces ($10^{10}$ or more), which naturally arise in many symbolic regression problems. We provide numerical demonstrations of the improved scalability of SIS compared to forward and backward feature selection in the Supporting Information. 
\end{remark}

\paragraph{Automated hyperparameter search} The number of expansion levels $l$ and the complexity constraint parameter $\lambda$ are two critical parameters. Thus, SyMANTIC sequentially loops over a finite number of these parameters, calling C$^2$-SISSO for every unique pair, i.e., 
\begin{align}
    (\mathcal{M}_{c,l}, \mathcal{L}_{c,l}, \mathcal{C}_{c,l}) = \text{C}^2\text{-SISSO}( \bs{y}, \bs{\Phi}_l, \lambda_c ), ~~~ \forall (c, l) \in \{1, \ldots, n_\text{comp} \} \times \{ 1, \ldots, n_\text{exp} \},
\end{align}
where we have used the subscripts $c$ and $l$ to denote the complexity constraint parameter and expansion level (with total number $n_\text{comp}$ and $n_\text{exp}$), respectively. The specific values of $\{ \lambda_1, \ldots, \lambda_{n_\text{comp}} \}$ are determined based on the range of feature complexity computed at a given level. Starting from $\lambda_1$ equal to the maximum complexity (no features dropped), we select subsequent $\lambda_c$ values so that they retain the $1 - (c-1)/n_\text{comp}$ fraction of the least complex features for $c > 1$. Since we loop over $c$ first, we can map $(c,l)$ to a global index $i = c + n_\text{comp}(l-1)$. Let $\mathcal{P}_0 = \emptyset$ be our initial approximation of the Pareto frontier between model loss and complexity. Then, we can update this approximation at every iteration
\begin{align}
    \mathcal{P}_i = \text{Update\_Pareto}( \mathcal{P}_{i-1},  \mathcal{M}_{c(i),l(i)}, \mathcal{L}_{c(i),l(i)}, \mathcal{C}_{c(i),l(i)} ), ~~ \forall i \in \{1, \ldots, n_\text{exp} n_\text{comp}\},
\end{align}
where $\text{Update\_Pareto}(\cdot)$ determines how the previous Pareto $\mathcal{P}_{i-1}$ should be updated given the newly obtained loss and complexity values after running C$^2$-SISSO at global iteration $i$. We can easily incorporate a stopping criteria if a loss threshold is met, meaning we can potentially terminate at some iteration $i < n_\text{exp} n_\text{comp}$. 

We highlight a key advantage of SyMANTIC is that it evaluates performance and complexity of at most a total of $n_\text{exp} n_\text{comp} \sum_{t=1}^T {tk \choose t}$ discrete models (symbolic expressions in the potentially infinite space $\mathcal{F}$). Therefore, the computational cost can be easily controlled through proper selection of $T$, $k$, $n_\text{comp}$, and $n_\text{exp}$. In our experience, default values of $T=3$, $k=20$, $n_\text{comp} = 4$, and $n_\text{exp} = 3$ yield a good balance between accuracy and computational time, as shown in our numerical experiments. Under these defaults, a maximum of 420240 models are tested, which is only a small fraction of all possible models -- the intelligent selection of which models to try via C$^2$-SISSO allows us to avoid wasting time testing many poor performing models (that are likely far from the Pareto front). Furthermore, by taking advantage of parallel computing, we can reduce the time to evaluate both the loss and complexity for all of these models on the order of seconds, as discussed in the next section.

\subsection{Software Implementation}
\label{subsec:software-impl}

We provide an efficient Python implementation of SyMANTIC that can be easily installed using the pip package manager using the following command: \texttt{pip install symantic}. The code is also available on Github: \texttt{https://github.com/PaulsonLab/SyMANTIC}. After installation, one can import the main \texttt{SymanticModel} class, create an instance of this class by passing relevant arguments, and then calling the \texttt{.fit()} method to return a dictionary populated with important outputs (such as the models along the approximate Pareto front and the ``best model'' corresponding to the utopia point of the approximate Pareto). A standard use case of the SyMANTIC code can be compactly written as follows:
\definecolor{LightGray}{gray}{0.95}
\begin{lstlisting}[language=Python]
# import SyMANTIC model class along with other useful packages
from symantic import SymanticModel
import numpy as np
import pandas as pd
# create dataframe composed of targets "y" and primary features "X"
data = np.column_stack((y, X))
df = pd.DataFrame(data)
# create model object to contruct full Pareto using default parameters
model = SymanticModel(data=df, pareto=True)
# run SyMANTIC algorithm to fit model and return dictionary "res"
res = model.fit()
# generate plot of Pareto front obtained during the fit process
model.plot_Pareto_front()
# extract symbolic model at utopia point and relevant metrics
model = res['utopia']['expression']
rmse = res['utopia']['rmse']
r2 = res['utopia']['r2']
complexity = res['utopia']['complexity']
\end{lstlisting}

Our \texttt{symantic} package is built on top of the PyTorch \cite{paszke2019pytorch} ecosystem, which offers two major advantages: (i) easy access to specialty compute hardware, e.g., graphics processing units (GPUs) and (ii) efficient implementation of an array of multivariate tensor operations that can be seamlessly integrated with the available hardware. A combination of these two advantages implies we can efficiently parallelize the most expensive operations in SyMANTIC. Specifically, C$^2$-SISSO involves fitting a potentially large number of (small) models as well as evaluating their structural complexity. For example, our implementation uses the \texttt{torch.linalg.lstsq} function to simultaneously compute a batch of solutions to the least squares problems to obtain the residuals $\bs{r}_t$ for all combination of models with $t$ terms. Note that, as $t$ increases, the batch size quickly grows such that keeping the maximum number of terms $T$ small greatly helps in keeping the computing requirements reasonably low. Larger $T$ will, e.g., make the batch size exceed the buffer size of the GPU, which will require a sequential processing of a maximum batch size. However, we have yet to run into this issue for any of our tested case studies. We also developed a similar approach to compute the structural complexity of all features using tensor operation when the matrix $\bs{\Phi}_l$ is being computed. This gives us a substantial improvement in performance compared to sequentially computing complexity using the SymPy \cite{meurer2017sympy}, as done in previous packages.  

We also incorporate dimensional analysis in \texttt{symantic} to improve efficiency by reducing the number of generated features during the feature expansion step of the SyMANTIC algorithm. This is achieved by defining units as symbols in SymPy, allowing us to track these units as operations are performed. Consequently, we prevent operations between inconsistent units. For example, both the addition $(+)$ and subtraction $(-)$ operations require that their inputs share the same units such that we can exclude generating new features for any combination of inputs with incompatible units. 

\subsection{Practical Considerations}

In this section, we discuss some practical considerations relevant to the effective application of SyMANTIC. Specifically, we discuss the role of hyperparameters, the possible model structures that can be found by SyMANTIC, and the impact of measurement noise and the distribution of the training data. In addition, we describe how SyMANTIC can be adapted to handle multiple outputs and learn models for dynamic systems.

\subsubsection{Hyperparameters in SyMANTIC}

The performance of SyMANTIC depends on several key hyperparameters, briefly mentioned in Section \ref{subsec:algorithm}. Below, we provide a summary for completeness. Unless otherwise stated, all numerical experiments in Section \ref{sec:case-studies} use the default value. While these defaults perform well across a variety of problems, systematic hyperparameter optimization may yield further improvements, which is outside the scope of this paper.

\paragraph{$\mathcal{O}$ (Operator Set)} The set of operators $\mathcal{O}$ controls the space of functions SyMANTIC can recover. The default choice is given in \eqref{eq:operator-set-default}. A well-chosen $\mathcal{O}$, informed by prior knowledge, can significantly enhance performance. Including too many operators may lead to exponential growth in the candidate descriptor space (as shown in \eqref{eq:growth-in-feature}). A key advantage of SyMANTIC’s efficiency is that it enables rapid experimentation with different choices of $\mathcal{O}$, which is particularly useful for real-world problems.

\paragraph{$n_\text{screen}$ and $\gamma$} These parameters control feature selection during expansion. $n_\text{screen}$ specifies the maximum number of features retained for regression (default: 20), and $\gamma$ is the mutual information threshold for feature viability (default: 0.1). Larger $n_\text{screen}$ increases the candidate model space but also computational cost.

\paragraph{$k$ (SIS Features)} This parameter determines the number of features retained in the SIS procedure, with a default of 20. Larger $k$ improves model quality but increases the computational requirements. For $k > 50$, solving the $\ell_0$ regression problem becomes infeasible, necessitating alternative approaches such as LASSO regression for higher values, which we plan to explore in future work.

\paragraph{$T$ (Model Terms)} $T$ sets the maximum number of terms in the final model, with a default of 3. Larger $T$ expands the complexity of the recovered models but complicates solving the $\ell_0$ regression problem. We recommend keeping $T \leq 5$ to balance model simplicity and computational feasibility. $T$ effectively acts as the ``regularization parameter'' in the constrained regression problem (i.e., the number of non-zero coefficients must be less than or equal to $T$ that is known be closely related to the penalty factor in penalty-based formulations).

\paragraph{$n_\text{comp}$ (Complexity Thresholds)} This specifies the number of complexity thresholds to test (default: 4). Preliminary experiments indicate limited sensitivity to this parameter, with linear scaling in computational cost.

\paragraph{$n_\text{exp}$ (Expansion Levels)} This parameter sets the maximum expansion depth (default: 3), which has a strong impact on both performance and computational cost due to exponential growth in the candidate descriptor space (see \eqref{eq:growth-in-feature}). To mitigate costs, we have incorporated an early-stopping threshold based on root mean squared error that can be specified by users. 
For initial (exploratory) experiments, setting $n_\text{exp} = 2$ can provide a significant reduction in computation time enabling rapid exploration of different operator sets $\mathcal{O}$.

\subsubsection{SyMANTIC's assumed model structure} 

SyMANTIC is currently designed to discover expressions that are sparse linear combinations of expanded features, generated by recursively applying all operators in $\mathcal{O}$ to combinations of the screened primary features. Specifically, the resulting model takes the form $\bs{\phi}_L^\top(\bs{x}) \bs{c}_L$, where $\| \bs{c}_L \|_0 \leq T$ (indicating at most $T$ non-zero terms) and $L \leq n_\text{exp}$ is the largest expansion level explored. While traditional sparse regression methods could be used to approximate solutions to such problems, the rapid growth in feature space size (potentially exceeding $10^6$ features) renders these approaches computationally prohibitive (or even infeasible). SyMANTIC addresses this challenge by offering a computationally efficient framework for tackling the problem while also providing a family of solutions via an approximate Pareto frontier. 

One current limitation is SyMANTIC's inability to identify unknown constants that appear nonlinearly in equations. For instance, a function such as $f(x) = \sin(2\pi x)$ can only be recovered if $2\pi$ is explicitly included in the operator set $\mathcal{O}$. There are existing methods, such as PySR \cite{cranmer2023interpretable}, that can handle more general model representations but at the cost of some other advantages that SyMANTIC brings. Identifying ways to modify SyMANTIC to work for more general nonlinear parametrizations and also potentially account for output constraints are interesting directions for future work.

\subsubsection{Impact of measurement noise} 

The current implementation of SyMANTIC uses a mean squared error (MSE) loss function, assuming additive i.i.d. Gaussian noise with zero mean and uniform variance across the measurements. For cases involving correlated Gaussian noise, a whitening transformation can be applied to the outputs before passing the data to SyMANTIC, as described in Section 2.4 of Muthyala et al. \cite{muthyala2024torchsisso}. In principle, other noise models could be accommodated by modifying the loss function. However, this would require adapting the $\ell_0$ regression step in PyTorch, which currently relies on batch calls to \texttt{torch.linalg.lstsq}, that is currently only applicable to MSE loss functions. Extending this to support custom loss functions is non-trivial (as one would need to build custom solutions for any specific form of $\ell(\cdot)$) and represents an interesting avenue for future work.

The level of noise in the measurements can significantly impact the quality of learned models. To address this, thorough validation and testing are essential during training, using methods such as held-out test data, $k$-fold cross-validation, or a combination of both approaches. The extent of testing is highly dependent on the application and should align with the use case for the model. 
For example, in safety-critical applications, users should carefully assess the model's robustness and remain cautious about extrapolating beyond the training regime, where performance is less predictable.

\subsubsection{Impact of the data distribution}

The distribution of the training data $\mathcal{D}$ is a critical factor in determining the quality of the final learned model, as with any supervised learning method. Similar to the effects of measurement noise, the influence of $\mathcal{D}$ on the learned model is challenging to predict \textit{a priori} and must be assessed empirically using validation and testing techniques. 

Our previous work \cite{muthyala2024torchsisso} illustrates this issue with an example where insufficient coverage of the input space leads to multiple model structures achieving comparable accuracy locally. In such cases, the problem becomes effectively ill-posed, and the best a supervised learning (or SR) algorithm can do is return the set of models that fit the data equally well. 
Fortunately, these situations can often be identified during testing. Systematic prediction errors, especially those occurring outside the range of the training data, highlight gaps in $\mathcal{D}$. Addressing this involves augmenting the training set with additional data from the problematic regions, followed by retraining the model to (hopefully) improve its generalizability.

\subsubsection{Handling multiple outputs} 

For systems with multiple output dimensions, one can apply SyMANTIC independently to each output. Specifically, given a dataset with $m$ outputs $\{y^{(1)}, y^{(2)}, \dots, y^{(m)}\}$, SyMANTIC can be used to solve $m$ separate single-output regression problems. For each output $y^{(i)}$, the model is learned as 
\begin{align}
   y^{(i)} \approx \bs{\phi}_{L^{(i)}}^\top(\bs{x}) \bs{c}_{L^{(i)}}^{(i)} ~~ \text{where} ~~  \|\bs{c}_L^{(i)}\|_0 \leq T,
\end{align}
where $L^{(i)} \leq n_\text{exp}$ is the largest expansion level explored for the $i^\text{th}$ output. 
This is straightforward approach that does not explicitly model possible correlations between the outputs. 
While effective for many problems, this may not fully leverage potential dependencies among the outputs, which could improve both interpretability and predictive performance in certain cases. Extending the SyMANTIC framework to directly account for multiple (especially high-dimensional) outputs is another interesting direction for future work. 

\subsubsection{Handling dynamical systems} 
\label{subsubsec:dynamics}

While our primary focus in Section \ref{subsec:prob-description} was on algebraic equations, SyMANTIC can also be applied to nonlinear dynamical systems of the form:
\begin{align} \label{eq:ode}
    \dot{\bs{z}}(t) = \bs{f}(\bs{z}(t)),
\end{align}
where $\bs{z}(t) = [z_1(t), z_2(t), \ldots, z_q(t)]^\top \in \mathbb{R}^q$ represents the system's state at time $t$, and $\bs{f} : \mathbb{R}^q \to \mathbb{R}^q$ specifies the dynamic constraints. Following a strategy similar to SINDy \cite{brunton2016discovering}, SyMANTIC can operate on a time-history of state measurements $\bs{z}(t)$ and their derivatives $\dot{\bs{z}}(t)$, collected at discrete time points $t_1, \ldots, t_r$. In this setup, $\{ \dot{\bs{z}}(t_1), \ldots, \dot{\bs{z}}(t_r) \}$ serve as the target data, while $\{ \bs{z}(t_1), \ldots, \bs{z}(t_r) \}$ are the input (primary feature) data. This framework can also be extended to handle more general functions $\bs{f}$ that depend on additional variables such as time $t$, parameters $\bs{\theta}$, or external forcing/control inputs $\bs{u}(t)$, as described in Section 3.3 of Brunton et al. \cite{brunton2016discovering}.

A key assumption of this type of approach is the availability of derivative information for the state. However, in many real-world scenarios, only state data for $\bs{z}(t)$ is accessible. In such cases, numerical differentiation techniques, such as total variation regularized differentiation \cite{chartrand2011numerical}, can estimate $\dot{\bs{z}}(t)$ while also filtering noise. The effectiveness of these methods depends on the density and quality of the available data. When measurements are sparse, numerical differentiation may produce unreliable estimates. To address this limitation, one can explore advanced methods like the Runge-Kutta variant of SINDy \cite{goyal2022discovery}, numerical integration techniques \cite{chen2018neural}, or nonlinear dynamic optimization approaches \cite{lejarza2022data}, which are better suited for sparse or noisy data. Integrating these ideas into SyMANTIC represents an exciting direction for future research.

Lastly, SyMANTIC can also be applied to discrete-time dynamical systems. In this case, the continuous-time equation \eqref{eq:ode} is replaced by a difference equation $\bs{z}_{k+1} = \bs{f}(\bs{z}_k)$ where $\bs{z}_k$ denotes the state at discrete time index $k$. If $\bs{z}_k$ alone does not fully represent the system's state (e.g., in time series modeling), lagged states can be incorporated, leading to a model of the form $\bs{z}_{k+1} = \bs{f}(\bs{z}_k, \bs{z}_{k-1}, \ldots, \bs{z}_{k-L})$ where $L \geq 0$ specifies the number of lags. Here, the targets are $\bs{z}_{k+1}$ and the inputs (primary features) are $\{ \bs{z}_k, \bs{z}_{k-1}, \ldots, \bs{z}_{k-L} \}$ at selected time indices $k$.

\section{Case Studies}
\label{sec:case-studies}

In this section, we demonstrate the effectiveness of SyMANTIC across a wide range of problems compare its performance to several popular SR methods. First, we consider a set of 20 synthetic and science-based benchmark equations wherein we provide a comprehensive comparison of SyMANTIC's performance to 5 open-source SR packages. Next, we compare the approximate Pareto frontiers found by SyMANTIC and the alternative SR packages on a problem with different explanations of varying complexity and a problem with varying levels of measurement noise. Then, we compare the predictive (test) performance of these different methods on a previously published competition dataset derived from an industrial chemical process. We then show how SyMANTIC can also be used for learning governing (differential) equations from data on the chaotic Lorenz system, and compare its performance on this task to SINDy (a popular nonlinear system identification approach). We then show another interesting use case for SyMANTIC in the context of molecular property modeling for battery applications. Lastly, we show how SyMANTIC benefits from GPU acceleration when scaling to larger tasks involving more data and more complex ground-truth equations. 

\subsection{Results and Comparisons for Benchmark Equations}

\paragraph{Datasets} First, we compare the performance of SyMANTIC to state-of-the-art open-source SR packages on benchmark regression tasks. In particular, we consider a total of 20 problems with data generated from ground-truth equations that are either synthetic in nature or originate from various fields of physics and engineering. The complete set of equations, and other relevant details such as the number of features, amount of training data, and noise level used for each equation, are summarized in Table \ref{tab:benchmark-equations}. Additional details such as the operator set and the data distributions are summarized in the Supporting Information. The first 5 synthetic equations (1-5) contain noise and randomly generated ``dummy features'' that do not influence the final output. The next 5 synthetic equations (6-10) contain ``deeper'' expression trees, requiring more levels of expansion to identify the ground truth. The subsequent 5 test equations (11-15) are taken directly from the EmpiricalBench dataset \cite{cranmer2023interpretable}, which was developed to be representative of the equation discovery task in science. Every equation in EmpiricalBench is a real expression discovered by a scientist from noisy, experimental data. The final 5 equations (16-20) are taken from the SRSD-Feynman dataset \cite{matsubara2022rethinking}. This data, originally proposed by Udrescu and Tegmark \cite{udrescu2020ai}, has been modified to ensure a realistic sampling range of values for the features and constants. 

\addtolength{\tabcolsep}{2pt}
\begin{table}[]
    \caption{Test equations that consist of a combination of synthetic problems, expressions from the EmpiricalBench dataset \cite{cranmer2023interpretable}, and the SRSD-Feynman dataset \cite{matsubara2022rethinking}. We provide the ground-truth equation, the number of features $D$, the number of training data points $n_\text{train}$, and the noise distribution if it is present. See the Supporting Information for additional details.}
    \centering
    \begin{tabular}{rlllllll}
    \hline
    \#  &Name  &Equation &$D$ &$n_\text{train}$ &Noise \\\hline
     1  &Synthetic\_1 &$y = \frac{10x_1}{x_2(x_3+x_4)}$ &5 &10 &$\mathcal{N}(0, 0.05)$  \\[2mm]
     2  &Synthetic\_2 &$y = 3\sqrt{x_1} + 2.1\sin(x_2) + 3$ &5 &10 &$\mathcal{N}(0, 0.05)$ \\[2mm]
     3  &Synthetic\_3 &$y = 2.5382 \cos(x_4) + x_1^2 - 0.5$ &5 &10 &$\mathcal{N}(0, 0.01)$ \\[2mm]
     4  &Synthetic\_4 &$y = \frac{x_1 + \exp(x_1)}{x_1^2 - x_2^2}$ &5 &10 &$\mathcal{N}(0, 0.01)$ \\[2mm]
     5  &Synthetic\_5 &$y = \sqrt{x_1^2 + x_2^2}$ &10 &10 &$\mathcal{N}(0, 0.05)$ \\[2mm]
     6  &Synthetic\_6 &$y = \exp(-x_1x_2) + \sin(x_1 x_3)$ &3 &10 &None \\[2mm]
     7  &Synthetic\_7 &$y = x_1^3 + 3 x_1 x_2^2 + 5 x_1 x_3^3$ &3 &10 &None \\[2mm]
     8  &Synthetic\_8 &$y = x_1^3 + x_1^2 + x_1$ &1 &10 &None \\[2mm]
     9  &Synthetic\_9 &$y = x_1^4 - x_1^3 + 0.5x_2^2 - x_2$ &2 &10 &None \\[2mm]
     10  &Synthetic\_10 &$y = \sin(x_1^2) \cos(x_1) - 2$ &10 &10 &None \\[2mm]
     11  &Hubble's law &$v=H_0 D$  &2 &10 &None \\[2mm]
     12  &Newton's law &$F = \frac{G m_1 m_2}{r^2}$  &4 &10 &None  \\[2mm]
     13  &Leavitt's law &$M = \alpha \log_{10}(P) + \delta$  &3 &10 &None  \\[2mm]
     14  &Ideal gas law &$P = \frac{nRT}{V}$  &4 &10 &None  \\[2mm]
     15  &Rydberg formula &$\frac{1}{\lambda}=R_H \left( \frac{1}{n_1^2} - \frac{1}{n_2^2} \right)$  &3 &10 &None  \\[2mm]
     16  &Distance &$d^2 = (x_2 - x_1)^2 + (x_3 - x_4)^2$ &4 &50 &None  \\[2mm]
     17  &Relativistic mass &$m^2 = \frac{m_0^2}{1 - \frac{v^2}{c^2}}$  &2 &50 &None  \\[2mm]
     18  &EM position &$x = \frac{qE}{m(\omega_1^2 - \omega_2^2)}$  &5 &50 &None  \\[2mm]
     19  &EM force &$F = q(E + Bv\sin(\theta))$  &5 &50 &None  \\[2mm]
     20  &Potential energy &$F = Gm_1m_2\left( \frac{1}{r_2} - \frac{1}{r_1} \right)$  &4 &50 &None  \\\hline
    \end{tabular}
    \label{tab:benchmark-equations}
\end{table}
\addtolength{\tabcolsep}{-3pt}

\paragraph{SR Methods} In addition to our proposed SyMANTIC method described in Section \ref{sec:symantic-method}, we consider the following 5 open-source SR packages:
\begin{itemize}
    \item \textbf{PySR} \cite{cranmer2023interpretable} is a recently proposed SR package that uses a custom evolutionary algorithm to optimize a loss function that accounts for both prediction error and complexity. It is optimized for performance with support for multi-threading and GPU acceleration. The original paper \cite{cranmer2023interpretable} reported that PySR outperforms all other tested algorithms, at least for scientific problems. We use the implementation available at \texttt{https://github.com/MilesCranmer/PySR}. We keep the default settings, except that we increased the number of iterations allowed.
    \item \textbf{PyOperon} is a Python wrapper to Operon \cite{burlacu2020operon}, which is a popular SR method that uses genetic programming to explore a hypothesis space of possible symbolic expressions. La Cava et al. \cite{la2021contemporary} recently performed a comprehensive study of 14 SR methods across a set of 252 problems and found that Operon was the most accurate method. We use the implementation available at \texttt{https://github.com/heal-research/pyoperon} with default settings.
    \item \textbf{gplearn} implements a traditional genetic programming method for SR and is a common benchmark method for comparison in the SR literature. We use the implementation at \texttt{https://github.com/trevorstephens/gplearn}, with default settings. Note that we did have to slightly modify certain functions to avoid potential issues such as numerical error and division by small numbers. 
    \item \textbf{DSO} \cite{petersen2021deep} is a deep learning-based framework for symbolic regression that leverages a recurrent neural network to generate distributions over mathematical expressions. It introduces a novel risk-seeking policy gradient method to train the network, optimizing it to produce expressions that better fit the data. DSO won first place in the real-world track of the 2022 SRBench Symbolic Regression Competition held at the GECCO 2022 conference. It also performed well on the synthetic track. We use the implementation at \texttt{https://github.com/dso-org/deep-symbolic-optimization} with default settings.
    \item \textbf{GP-GOMEA} \cite{virgolin2021improving} is a genetic programming-based SR method. It employs linkage learning to identify dependencies between variables, allowing for more efficient exploration of the search space. GP-GOMEA was found to be among the top-performing algorithms in the comprehensive study done by La Cava et al. \cite{la2021contemporary} where it was slightly less accurate than Operon but produced less complex expressions. We used the implementation at \texttt{https://github.com/marcovirgolin/GP-GOMEA} with default settings.
\end{itemize}
Note that all experiments were conducted on a computing cluster with two nodes, each equipped with an Intel Xeon Gold 6444Y processor with 16 cores and 512 GB of DDR4 RAM. We imposed a relatively aggressive time limit of 5 minutes per problem for all algorithms, significantly shorter than the 8-hour limit used in the standard SRBench case \cite{la2021contemporary}. The rationale for this was twofold: (i) a longer time limit would require substantial computational resources, and (ii) SyMANTIC is designed to be both fast and accurate, making performance in resource-constrained settings a key focus of this work.

\paragraph{Performance Metrics} We use three key metrics to evaluate the performance of tested SR methods. First, we calculate the percentage of ground-truth equations recovered. For example, an algorithm that correctly discovers all 20 equations in Table \ref{tab:benchmark-equations} would receive a score of 100\%. To avoid issues with small errors in the numerical constants, we manually checked the equations returned by each algorithm (automated checks using SymPy \cite{meurer2017sympy} often failed to produce correct results). Second, we evaluate the structural complexity of the model, as defined in \eqref{eq:structural-complexity}. This information-theoretic metric (measured in bits) accounts for both the number of unique inputs and mathematical operators as well as the number of times they are used. Third, we evaluate the ``training time'' that measures how long it required for the method to terminate.
Due to randomness in the data generation process and the genetic programming-based methods, we perform 5 repeated trials from different random seeds for every algorithm. We report the performance metrics averaged across these trials.

\begin{figure}[tb]
\centering
\includegraphics[width=1.0\textwidth]{ 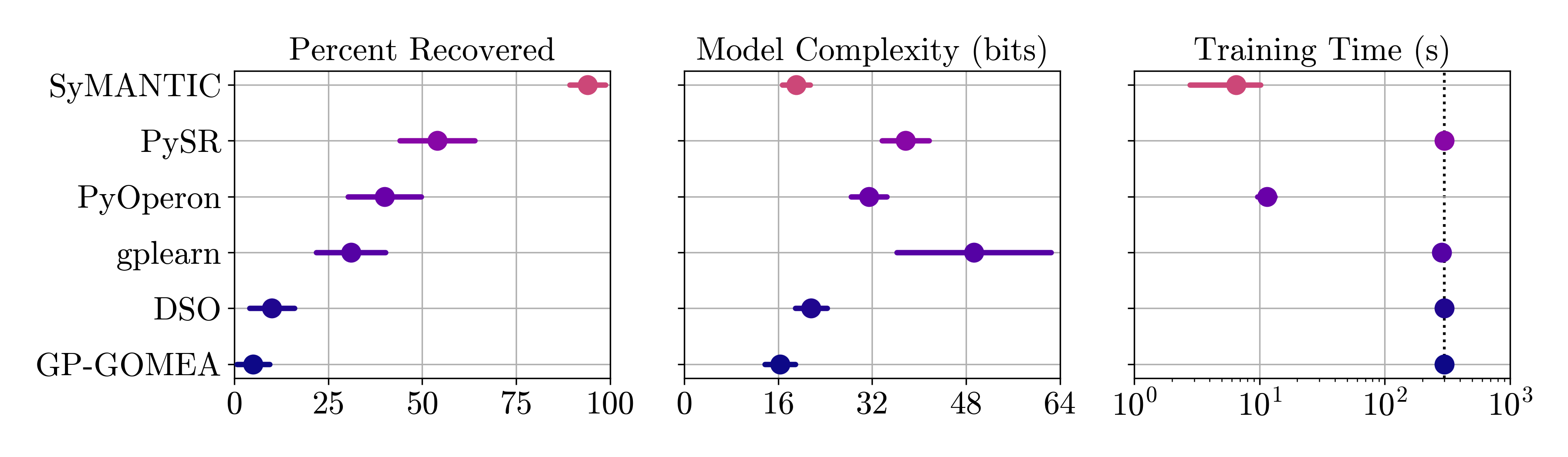}
\caption{Results on the full set of test equations in Table \ref{tab:benchmark-equations} for all algorithms. Left shows the percent of ground-truth equations recovered, middle shows the median model complexity across all test problems, and right shows the median training time across all test problems. Points indicate the mean of the metrics across the 5 replicates while bars show the estimated 95\% confidence intervals. The dotted line in the training time plot represents the 5 minute time limit imposed per problem on all methods.}
\label{fig:all-alg-comparison}
\end{figure}

\paragraph{Results} The results for the 20-problem benchmark task are summarized in Figure \ref{fig:all-alg-comparison}, which presents the percentage of recovery, median model complexity, and median training/solution time across all problems. SyMANTIC demonstrates superior performance, exactly recovering over 95\% of the test equations (averaged across five replicates), significantly outperforming the next best method, PySR, which recovers approximately 50\%. Again we reiterate that half of these test equations are real-world scientific equations, historically discovered by humans and previously shown to be challenging to identify from data.
SyMANTIC also produces significantly less complex models compared to most tested packages, underscoring its ability to balance accuracy and simplicity. Additionally, SyMANTIC has the shortest training time among all methods, with a median value of just under 10 seconds. This efficiency is attributed to the Python implementation described in Section \ref{subsec:software-impl}. In contrast, four other methods (PySR, gplearn, DSO, GP-GOMEA) consistently reached the 5-minute time limit on most problems. While these methods are likely to perform better with a larger computational budget, the challenge remains that most SR methods do not provide a clear indication of the required budget \textit{a priori}. Moreover, it is unclear how to allocate available computational resources -- whether to restart with a new random seed, adjust algorithm hyperparameters, or continue from the best solution found. SyMANTIC's fast convergence allows for more frequent testing of different configurations (e.g., changing the operator set), accelerating the overall discovery process.

\subsection{Pareto Front Comparison}

Next, we consider the relativistic momentum equation
\begin{align} \label{eq:rel-mom}
    p = \frac{mv}{\sqrt{1 - \frac{v^2}{c^2}}},
\end{align}
where $p$ denotes the relativistic momentum of an object, $m$ denotes the rest mass of the object, $v$ denotes the velocity of the object, and $c$ is the speed of light in a vacuum. We treat $\{ m, v, c \}$ as features with 50 training datapoints generated from the following distribution $m \sim \mathcal{U}_\text{log}(10^{-2}, 10^0)$, $v \sim \mathcal{U}_\text{log}(10^5, 10^7)$, and $c = 2.998 \times 10^8$ based on the SRSD-Feynman dataset \cite{matsubara2022rethinking}. We select this problem because, in addition to the ground-truth equation, there is the classical equation $p = mv$ that can be viewed as an approximate version of \eqref{eq:rel-mom} that neglects the Lorenz factor $\gamma = \frac{1}{\sqrt{1 - \frac{v^2}{c^2}}}$, which approaches 1 as $v$ gets smaller. Ideally, an SR method could simultaneously discover both theories from one dataset. 

\begin{figure}[tb]
\centering
\includegraphics[width=0.8\textwidth]{ 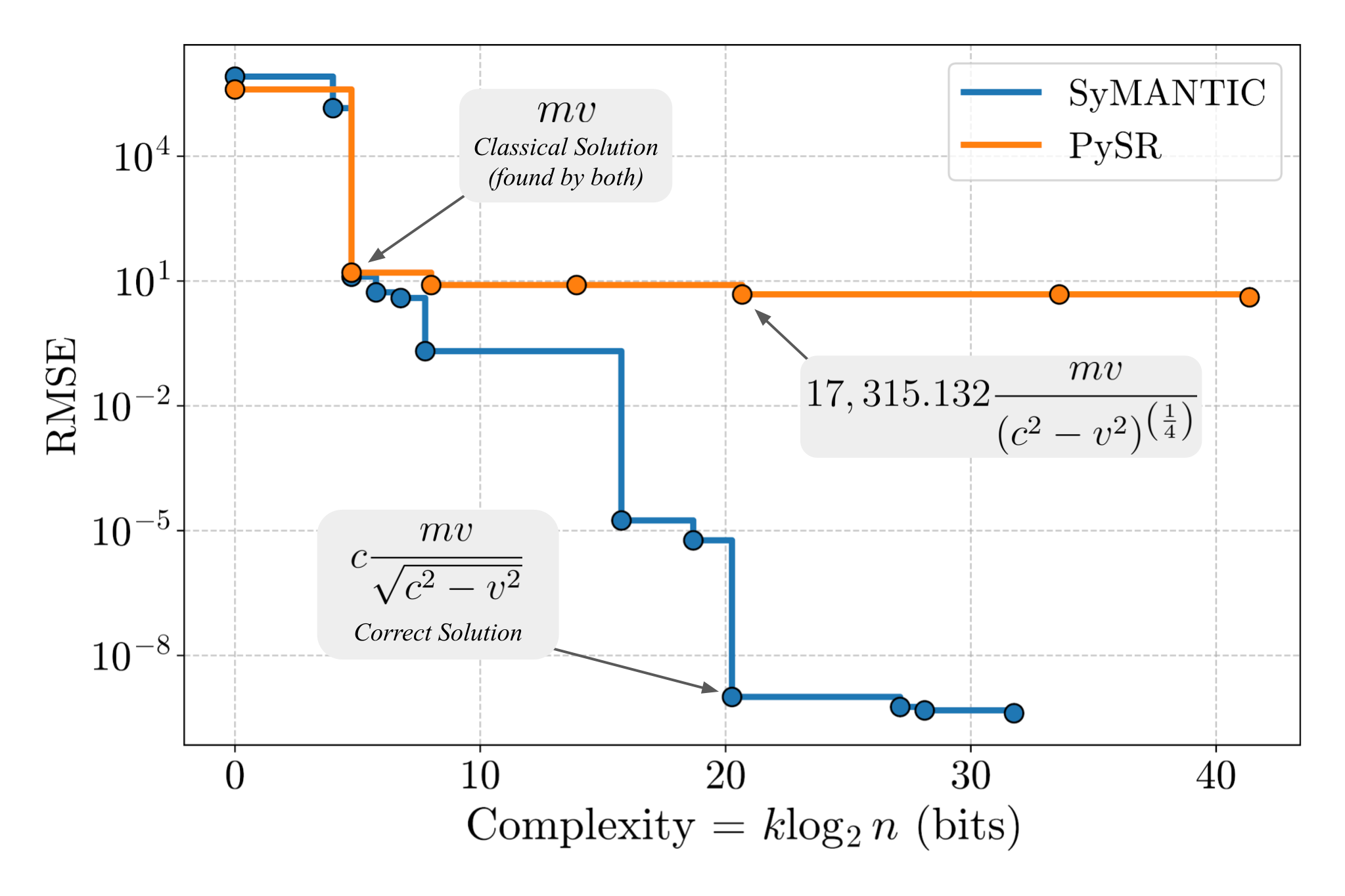}
\caption{Approximate Pareto fronts between root mean squared error (RMSE) and structural complexity found by SyMANTIC and PySR on the relativistic momentum problem.}
\label{fig:pareto-results}
\end{figure}

The approximate Pareto fronts of accuracy (RMSE) versus complexity found by SyMANTIC and PySR are shown in Figure \ref{fig:pareto-results}. We only compare directly to PySR, as it was the next best method in the comprehensive analysis done in the previous section (and it is one of the few methods that also returns an approximate Pareto front). Interestingly, both methods identify the classical solution that, as expected, provides a nice balance between accuracy and simplicity; however, only SyMANTIC successfully discovers the ground-truth equation. Additionally, SyMANTIC also ends up finding a few equations with slightly better RMSE, though these are clearly overfit to the finite amount of training data, as indicated by their higher complexity. This latter result highlights the importance of approximating the full Pareto front, which enhances robustness to noise and ``bad'' data (potentially with outliers).



\subsection{Comparison under Varying Noise Levels}

We now examine the robustness of SyMANTIC in the presence of challenging noise characteristics by incorporating heteroscedastic noise into the measurements. Specifically, we consider the Rydberg formula (Table \ref{tab:benchmark-equations}, \#15) under the same conditions described in the Supporting Information, with the following modification: the measurements now have the form $y_i = f(\bs{x}_i) + \epsilon_i$ where $\epsilon_i \sim \mathcal{N}(0, (v f(\bs{x}_i))^2)$ represents input-dependent noise with variance proportional to the target value and $v$ is a parameter controlling the noise magnitude. Here, $v$ can be interpreted as the percentage of noise added relative to the target's true value.

\begin{figure}[tb]
\centering
\includegraphics[width=0.8\textwidth]{ 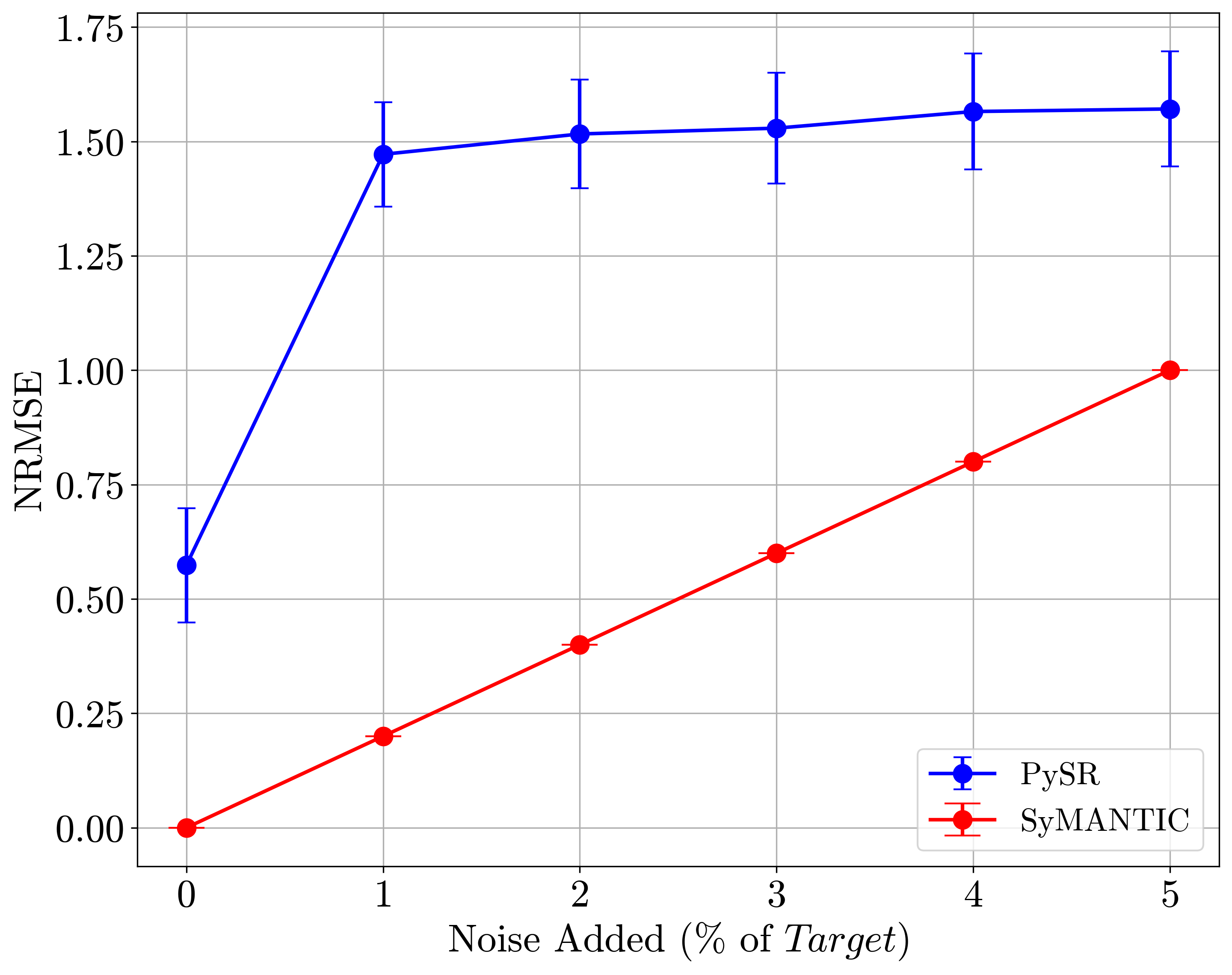}
\caption{Normalized root mean squared error (NRMSE) versus the percentage of noise added to the measured target values (referenced as parameter $v$ in the text) for the Rydberg formula for 10 independent replicates for SyMANTIC and PySR.}
\label{fig:noise-study}
\end{figure}

To evaluate the performance of SyMANTIC under these noisy conditions, we compute the normalized RMSE (NRMSE), defined as RMSE/($y_\text{max} - y_\text{min}$) where $y_\text{min}$ and $y_\text{max}$ denote the minimum and maximum target values, respectively. Figure \ref{fig:noise-study} compares the NRMSE over 10 independent replicates for SyMANTIC and PySR as a function of the noise parameter $v$. For PySR, we imposed a larger 10-minute time limit per run, which significantly exceeds the runtime required for individual SyMANTIC runs.
The results clearly demonstrate the superior noise resilience of SyMANTIC. Across all noise levels, SyMANTIC achieves substantially lower NRMSE compared to PySR, indicating more accurate recovery of the underlying model. As anticipated, the error increases for both methods as the noise parameter $v$ grows. However, SyMANTIC exhibits a much more consistent and predictable trend in error escalation, whereas PySR's performance shows increased variability. 

\subsection{Dow Chemical Industrial Competition Dataset}

To further highlight the power of SyMANTIC on challenging scientific problems, we next explore the so-called ``chemical 2 (competition)'' dataset derived from continuous processes at Dow Chemical \cite{kordon2008evolutionary}. This dataset has a target variable that was measured by collecting noisy and expensive lab data on the chemical composition of a valuable end-product. It also has 57 input variables derived from relatively inexpensive process measurements such as temperatures, pressures, and flowrates. This dataset was used in the SR competition at EvoStar conference 2010 and can be downloaded from their website \cite{symreg_competition_2012} (we have also uploaded a copy of the data in the Github page associated with this paper). We keep the same settings as that used in the competition, consisting of a set of 747 training points and 319 test points. Since the feature set is quite large, we slightly modified the MI screening procedure in SyMANTIC to keep features whose MI values were in the top 50\%. The train/test results for SyMANTIC and PySR are reported in Table \ref{tab:dow-results}. Again, we see that SyMANTIC has lower RMSE and higher $R^2$ on both the train and test set. Specifically, SyMANTIC achieves a test $R^2$ of 0.78 compared to 0.45 for PySR, implying it has learned an improved model that better generalizes beyond the training dataset. It is also worth noting that SyMANTIC achieves a lower test RMSE than other packages such as HeuristicLab and Operon reported in a recent study on this data \cite{radwan2024comparison}, even though they take considerably more time (e.g., HeuristicLab takes roughly 50 minutes to find an equation with a test RMSE of $\sim 0.154$ while SyMANTIC achieves a test RMSE of $\sim 0.026$ in less than 1 minute)\footnote{Note that the train/test sets reported in Radawan et al.\cite{radwan2024comparison} slightly differ from the original competition dataset. We also had to convert from normalized mean squared error (NMSE) to RMSE by $\text{RMSE} = \sqrt{\sigma^2_y\text{NMSE}}$ where $\sigma^2_y$ denotes the population variance of the target dataset.}.



\addtolength{\tabcolsep}{2pt}
\begin{table}[tb]
    \caption{Tabulated training and test performance of SyMANTIC and PySR on the Dow Chemical competition dataset.}
    \centering
    \begin{tabular}{lcccc}
    \hline
    Method & RMSE (train) &RMSE (test) &$R^2$ (train) &$R^2$ (test) \\\hline
    SyMANTIC & 0.0253 &0.0255 &0.797 &0.781 \\
    PySR     & 0.0628 &0.0637 &0.499 &0.453 \\\hline
    \end{tabular}
    \label{tab:dow-results}
\end{table}
\addtolength{\tabcolsep}{-3pt}

\subsection{Lorenz System with Chaotic Dynamics}

To further showcase the versatility of SyMANTIC, we next consider the discovery of dynamic governing equations from data. Specifically, we consider the chaotic Lorenz system \cite{lorenz1963deterministic}
\begin{subequations}
\begin{align}
    \dot{x} &= \sigma(y - x), \\
    \dot{y} &= x(\rho - z) - y, \\
    \dot{z} &= xy - \beta z, 
\end{align}
\end{subequations}
that describes the properties of a two-dimensional fluid layer uniformly warmed from below and cooled from above. Specifically, $x$ is a state variable proportional to the rate of convection, $y$ is a state variable proportional to the horizontal temperature variation, $z$ is a state variable proportional to the vertical temperature variation, and $\sigma$, $\rho$, and $\beta$ are constants.

The problem of learning these equations from measurements of the states $(x(t), y(t), z(t))$ and their derivatives $(\dot{x}(t), \dot{y}(t), \dot{z}(t))$ at specific time points $t \in \mathcal{T}$ was considered in original SINDy paper \cite{brunton2016discovering}. We use a similar setup to the original paper based on the implementation in the PySINDy \cite{de2020pysindy} package -- we use the same constants $\sigma = 10$, $\rho = 28$, $\beta = 8/3$ and initial conditions $x(0) = -8$, $y(0) = 8$, $z(0) = 27$, which are known to produce fairly chaotic behavior. We assume the measurements are noise-free, however, we assume the states are measured at only 5 time points randomly selected in the interval $[0, 100]$. This is considerably less data than that used in the original SINDy paper (factor of 4 fewer time points). 

\begin{figure}[tb]
\centering
\includegraphics[width=1.0\textwidth]{ 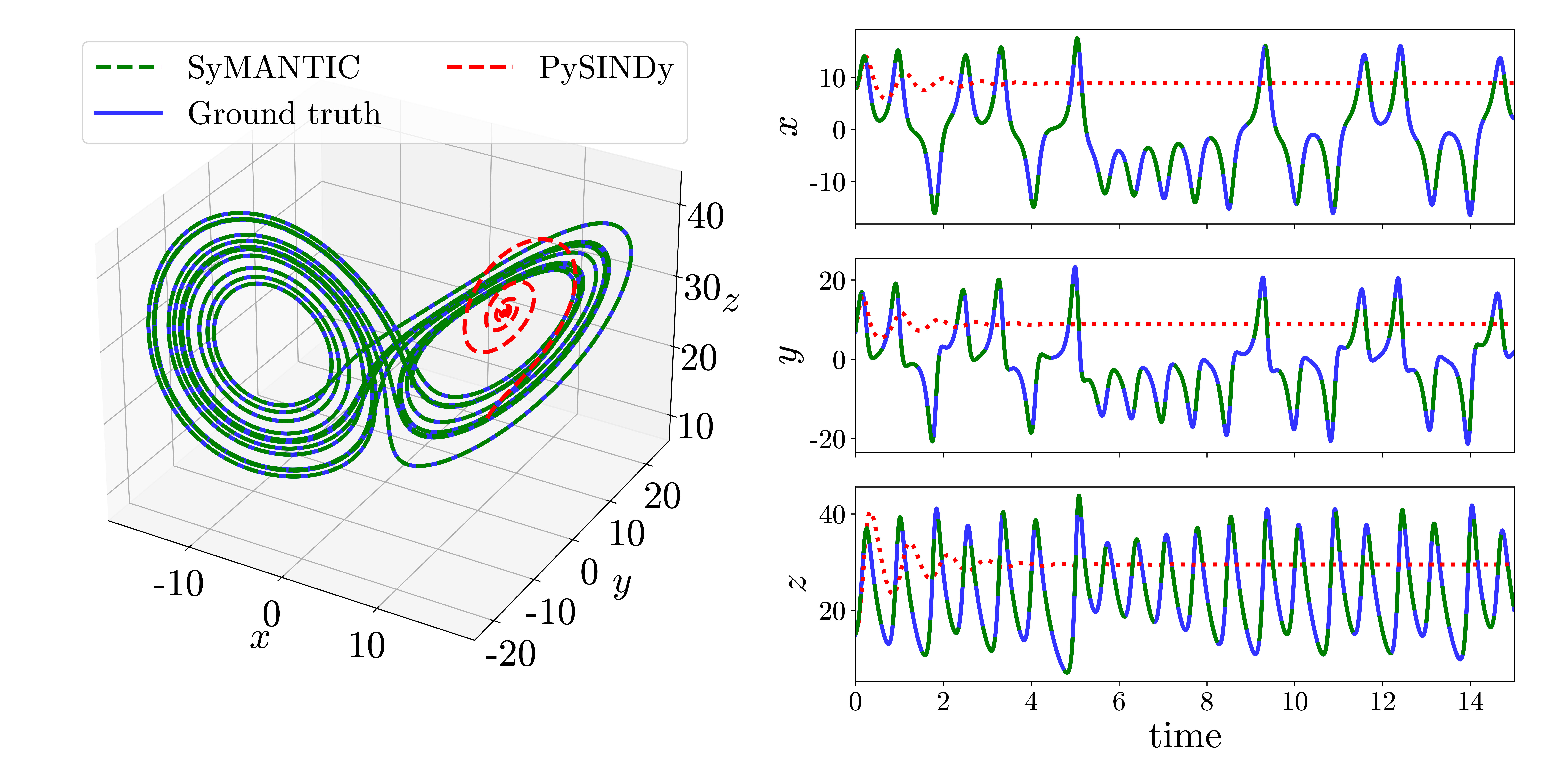}
\caption{Comparison of the predicted dynamic evolution of the chaotic Lorenz system based on the governing equations learned by SyMANTIC and PySINDy from training data. Left shows the phase plot of SyMANTIC (dashed green), PySINDy (dashed red), and ground truth (solid blue) over the time interval $[0, 15]$. Right shows the individual states $x(t)$, $y(t)$, and $z(t)$ over the same time interval. SyMANTIC is able to accurately capture the chaotic behavior present in the system from very sparse data, while PySINDy produces predictions that quickly diverge frome the ground truth.}
\label{fig:lorenz-results}
\end{figure}

The prediction results obtained by finely integrating the discovered equations over the interval of $[0, 15]$ with SyMANTIC and PySINDy (using the recommended default settings) are shown in Figure \ref{fig:lorenz-results}. We see that SyMANTIC identifies near-perfect governing ordinary differential equations that accurately capture the complete state evolution $(x, y, z)$ over time. The predictions made by PySINDy, on the other hand, quickly begin to deviate from the ground truth and in fact converge to a non-existent steady state (completely missing the chaotic behavior). We do note that PySINDy does perform well on this system after $\sim 10$ time points worth of measurements are available (see the Supporting Information for these and additional comparisons to a more recent variant, Ensemble-SINDy \cite{fasel2022ensemble}), however, this highlights the strong robustness of SyMANTIC even in the very sparse/low-data regime.

\subsection{Discovery of Molecular Property Models}

As a final case study, we consider a molecular property prediction problem adapted from our recent work \cite{park2024sparkle}. Specifically, we consider the prediction of redox potential, which is an important property for organic electrode materials (OEMs) that are promising (potentially cheap and sustainable) components for energy storage in rechargeable batteries. Although redox potential can be predicted using density functional theory (DFT), the computational cost of DFT calculations is large, making it effectively impossible to scale to large design spaces with millions of OEM candidates. We use the same training set of 115 paraquinones developed in our past work \cite{park2024sparkle}, and consider a test set of 1000 randomly selected quinone molecules from a previous study by Tabor et al. \cite{tabor2019mapping}. Unlike our previous work, we utilize a new set of molecular descriptors, primarily the open-source PaDEL set \cite{yap2011padel}, which converts the SMILES representation of a molecule into 1,444 input features. Due to the very high-dimensional input space, we modify the MI screening procedure in SyMANTIC to only keep features in the top 1\% of the MI values. 

To further highlight the advantages of SR methods over traditional machine learning approaches, we compare SyMANTIC with three well-established methods: Least Absolute Shrinkage and Selection Operator (LASSO), Random Forest (RF), and Neural Networks (NNs). LASSO is a standard $\ell_1$ sparse regression method, RF is an ensemble learning method based on decision trees, and NNs consist of layers of linear transformations combined with nonlinear activation functions. We implemented LASSO and RF using the scikit-learn package \cite{pedregosa2011scikit} and optimized their hyperparameters through 5-fold cross-validation. For the NN model, we used PyTorch \cite{paszke2019pytorch} and trained it with the Adam optimizer. Hyperparameter tuning for the NN, including the number of hidden layers, nodes per layer, activation function, and learning rate, was performed using the Optuna framework \cite{akiba2019optuna}, with 20\% of the training data randomly held out as a validation set.

Table \ref{tab:materials-results} presents the train/test results for SyMANTIC, PySR, LASSO, RF, and NN. While all methods perform well on the training set ($R^2 > 0.9$), only SyMANTIC and PySR maintain strong performance on the test set. SyMANTIC achieves the highest test $R^2$ (0.88), significantly outperforming PySR (0.70). Both LASSO and NN produce negative test $R^2$ values, indicating severe overfitting to the training data. This outcome is expected, as the small training set was chosen specifically to highlight the extrapolation capabilities of SR methods in low-data scenarios. It is noteworthy how well SR methods, particularly SyMANTIC, perform in this real-world setting, where it is unclear \textit{a priori} if a simple symbolic expression can effectively capture the underlying behavior. In addition to its robustness against overfitting, SyMANTIC has a faster training time than all methods except LASSO, identifying the best model in under 10 seconds.

\addtolength{\tabcolsep}{1pt}
\begin{table}[tb]
    \caption{Tabulated training and test performance of SyMANTIC, PySR, and traditional machine learning methods on the molecular property dataset (redox potential of quinone molecules). The LASSO and random forest (RF) models were trained using scikit-learn package \cite{pedregosa2011scikit}, with hyperparameters found by 5-fold cross validation. The neural newtork (NN) model was trained using PyTorch \cite{paszke2019pytorch}, with hyperparameters optimized using Optuna \cite{akiba2019optuna}.}
    \centering
    \begin{tabular}{lccccc}
    \hline
    Method & RMSE (train) &RMSE (test) &$R^2$ (train) &$R^2$ (test) &Training time (s) \\\hline
    SyMANTIC &0.0297 &0.071 &0.99 &0.88 &9.21 \\
    PySR     &0.0331 &0.115 &0.98 &0.70 &300 \\
    LASSO    &0.0106 &0.556 &0.99 &-6.05  &5.98 \\
    RF       &0.0615 &0.136 &0.94 &0.58  &70.00 \\
    NN       &0.0075 &6.214 &0.99 &-881.40  &134.89  \\\hline
    \end{tabular}
    \label{tab:materials-results}
\end{table}
\addtolength{\tabcolsep}{-1pt}

\subsection{GPU Hardware Acceleration}

As mentioned in Section \ref{subsec:software-impl}, a key advantage of our PyTorch-based implementation of SyMANTIC is its compatibility with various types of accelerated computer hardware, including multi-core CPUs and GPUs. For the benchmarking studies presented earlier, we ran all experiments on multi-core CPU hardware to ensure a fair comparison between methods, as the majority of alternative methods do not provide GPU support.

\begin{figure}[tb]
\centering
\includegraphics[width=0.8\textwidth]{ 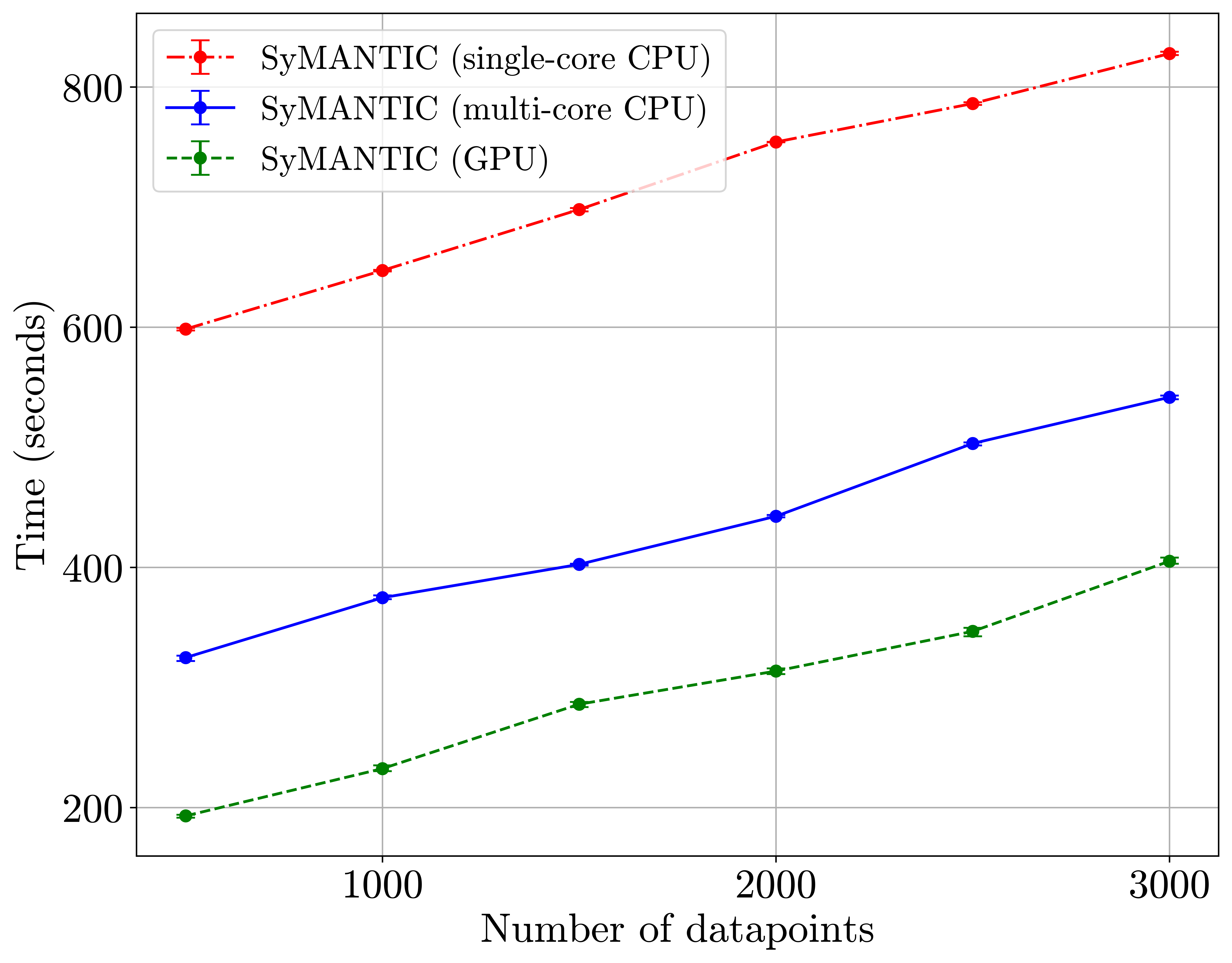}
\caption{Computational time versus number of training datapoints for running SyMANTIC on single-core CPU, multi-core CPU, and GPU hardware for a set of four term models.}
\label{fig:gpu-hardware}
\end{figure}

In this section, we highlight the potential runtime benefits of using GPUs for symbolic regression, particularly for more computationally intensive problems. We consider the following three ground-truth equations:
\begin{align*}
    & x_1^4 + x_2^3 + x_3^2 + x_4, \\
    & x_1^5 + x_2^4 + x_3^3 + x_4, \\
    & x_1^3 + x_2^2 + x_3 + \sin(x_4), 
\end{align*}
each of which consists of four terms. We updated the hyperparameter $T = 4$ accordingly. Figure \ref{fig:gpu-hardware} compares the computational time required to learn these models from randomly generated data (with inputs sampled uniformly over the range $[1,5]$) for three hardware configurations: single-core CPU, multi-core CPU, and GPU. The number of training data points varies from 500 to 3000.
These tests were run on an NVIDIA A100 GPU with 40 GB of RAM, alongside the same computer cluster used in earlier experiments, where CPU memory was capped at 40 GB of RAM for consistency. Error bars in Figure \ref{fig:gpu-hardware} reflect the minimum and maximum times observed across the three models. As expected, training time increases with problem complexity, driven by the growth in both $T$ and the number of datapoints. However, GPU acceleration demonstrates a clear advantage, consistently reducing training time by a factor of 2 to 3 in all cases.
Further improvements can be expected when running the method on multiple GPUs simultaneously by leveraging PyTorch's Distributed Data Parallel (DDP) functionality. This scalability makes SyMANTIC well-suited for tackling even larger and more complex symbolic regression problems. 

\section{Conclusion}
\label{sec:conclusion}

In this paper, we introduce SyMANTIC, a symbolic regression (SR) method designed to discover simple, interpretable mathematical expressions from data. SyMANTIC overcomes key challenges of existing SR approaches by offering low computational cost, robust handling of high-dimensional inputs and noisy data, and the ability to balance accuracy and simplicity. Instead of returning a single most accurate expression, SyMANTIC provides a set of (approximate) Pareto-optimal expressions that trade off between accuracy and simplicity. To enable this tradeoff, we propose an information-theoretic measure of structural complexity that accounts for both the number and frequency of unique inputs and operators. While constructing the exact Pareto front is generally infeasible, SyMANTIC intelligently searches a finite set of expressions likely near the Pareto front by building on the recently proposed SISSO framework. In particular, we develop a complexity-constrained variant (C$^2$-SISSO) that accounts for feature complexity during the search. To further enhance performance, we include an automated hyperparameter search that sequentially improves the approximate Pareto front at every iteration. SyMANTIC is released as an easy-to-use Python package built on the PyTorch ecosystem, enabling efficient data handling and GPU acceleration.

We extensively test SyMANTIC on a diverse set of problems, including synthetic examples, challenging scientific expressions, real-world chemical and material property prediction, and chaotic dynamical system identification, comparing its performance to several state-of-the-art SR methods. SyMANTIC consistently outperforms these alternatives, recovering over 95\% of ground-truth equations from benchmark SR tasks, while the next best method recovers just over 50\%. In addition to its superior accuracy, SyMANTIC has the shortest training time due to its efficient, easily parallelizable implementation. We also demonstrate SyMANTIC's ability to accurately learn the equations governing the chaotic Lorenz system from state derivative measurements at just five time points, outperforming the widely-used SINDy method. Finally, in a material property prediction problem relevant to sustainable battery materials, SyMANTIC learns more accurate, generalizable models than existing SR and traditional machine learning approaches, even with limited training data.

There are several promising directions for future work. One is applying SyMANTIC to a larger set of benchmark and real-world problems to identify any specific types of expressions with which it may struggle. Another is exploring alternative loss functions that could better handle outliers and non-Gaussian noise in the data. Additionally, it would be valuable to investigate hybrid approaches that combine the sparse linear regression framework that underpins SyMANTIC with genetic programming-based search. Specifically, a limitation of SyMANTIC is that it only explores expressions up to a maximum depth, while genetic programming can explore deeper expressions. However, existing genetic programming methods end up testing many low-quality expressions, making the search process inefficient. SyMANTIC's targeted search is a key reason for its success in the case studies presented here, and a hybrid approach could potentially combine the strengths of both methods.
\begin{suppinfo}
The Supporting Information is available free of charge at http://pubs.acs.org/. 
\begin{itemize}
    \item Additional details on the benchmark symbolic regression tasks including input features, data distribution, and operator set for each problem; a numerical comparison of SIS to more traditional forward/backward feature selection methods on an example problem; and additional comparisons of SyMANTIC to PySINDy and Ensemble-SINDy given more data and varying amount of noise in the observations (PDF).
\end{itemize}
\end{suppinfo}

\bibliography{references}



\section*{Supplementary material}

# Contents

# S1  Details on Synthetic Benchmark Problems

In this study, we generate synthetic datasets to evaluate the performance of our models across a variety of mathematical relationships and complexities. Each dataset represents a different case with a unique underlying function. Below, we provide detailed information on how the data for each case is generated, including the mathematical functions used, input variable ranges, noise levels, and any additional pertinent details.

- **Sample Size**: Each dataset consists of 10 samples.

- **Input Variables** ($X$): For each case, a set of input variables $x_1, x_2, \ldots, x_n$ is generated. The number of variables $n$ varies per case.

- **Variable Distribution**: Each input variable $x_i$ is independently sampled from a uniform distribution over the interval $[1, 5]$:

$$x_i \sim \mathcal{U}(1, 5)$$

## S1.1  Case 1

- **Mathematical Function**:

$$y = 10 \left( \frac{x_1}{x_2(x_3 + x_4)} \right) + 2.5 + \epsilon$$

- **Input Variables**:

  - $x_1, x_2, x_3, x_4, x_5$

  - Only $x_1, x_2, x_3, x_4$ are used in $y$; $x_5$ is generated but not used.

- **Noise**:

- $\epsilon \sim \mathcal{N}(0, 0.05)$

- **Operators Used in Symbolic Regression**:

  - Division (/)

  - Addition (+)

- **Termination Criteria (Metrics) for SyMANTIC**:

  - Root Mean Square Error (RMSE) threshold: 0.04

  - Coefficient of Determination ($R^2$) threshold: 0.99

## S1.2 Case 2

- **Mathematical Function**:

$$y = 3\sqrt{x_1} + 2.1\sin(x_2) + 3 + \epsilon$$

- **Input Variables**:

  - $x_1, x_2, x_3, x_4, x_5$

  - Only $x_1$ and $x_2$ are used in $y$; $x_3, x_4, x_5$ are generated but not used.

- **Noise**:

  - $\epsilon \sim \mathcal{N}(0, 0.05)$

- **Operators Used in Symbolic Regression**:

  - Sine function (sin)

  - Square root function ($\sqrt{\ }$)

- **Termination Criteria (Metrics) for SyMANTIC**:

- RMSE threshold: 0.04
- $R^2$ threshold: 0.99

### S1.3 Case 3

- **Mathematical Function**:

$$y = 2.5382 \cos(x_4) + x_1^2 - 0.5 + \epsilon$$

- **Input Variables**:
  - $x_1, x_2, x_3, x_4, x_5$
  - Only $x_1$ and $x_4$ are used in $y$; $x_2, x_3, x_5$ are generated but not used.

- **Noise**:
  - $\epsilon \sim \mathcal{N}(0, 0.01)$

- **Operators Used in Symbolic Regression**:
  - Cosine function (cos)
  - Power function (square, $x^2$)

- **Termination Criteria (Metrics) for SyMANTIC**:
  - RMSE threshold: 0.01
  - $R^2$ threshold: 0.99

### S1.4 Case 4

- **Mathematical Function**:

$$y = \frac{x_1 + e^{x_1}}{x_1^2 - x_2^2} + \epsilon$$

- **Input Variables**:

  - $x_1, x_2$

  - Both variables are used in $y$.

- **Noise**:

  - $\epsilon \sim \mathcal{N}(0, 0.01)$

- **Operators Used in Symbolic Regression**:

  - Power function (square, $x^2$)

  - Addition (+)

  - Division (/)

  - Exponential function (exp)

  - Subtraction (−)

- **Termination Criteria (Metrics) for SyMANTIC**:

  - Default criteria used.

## S1.5 Case 5

- **Mathematical Function**:
$$y = \sqrt{x_1^2 + x_2^2}$$

- **Input Variables**:

  - $x_1, x_2, x_3, \ldots, x_{10}$

  - Only $x_1$ and $x_2$ are used in $y$; the other variables are generated but not used.

- **Noise**:

- $\epsilon \sim \mathcal{N}(0, 0.05)$

- **Operators Used in Symbolic Regression**:

  - Power function (square, $x^2$)
  - Square root function ($\sqrt{\ }$)
  - Addition ($+$)

- **Termination Criteria (Metrics) for SyMANTIC**:

  - RMSE threshold: 0.001
  - $R^2$ threshold: 1.0

## S1.6  Case 6

- **Mathematical Function**:

$$y = e^{-x_1 x_2} + \sin(x_1 x_3)$$

- **Input Variables**:

  - $x_1, x_2, x_3$
  - All variables are used in $y$.

- **Noise**:

  - No noise is added ($\epsilon = 0$).

- **Details**:

  - The function combines an exponential decay term and a sine function involving products of variables.
  - This introduces complex variable interactions and nonlinearity.

- **Operators Used in Symbolic Regression**:

    - Multiplication ($\times$)

    - Exponential function (exp)

    - Sine function (sin)

    - Inverse function ($x^{-1}$)

- **Termination Criteria (Metrics) for SyMANTIC**:

    - RMSE threshold: 0.001

    - $R^2$ threshold: 1.0

## S1.7 Case 7

- **Mathematical Function**:

$$y = x_1^3 + 3x_1x_2^2 + 5x_1x_3^3$$

- **Input Variables**:

    - $x_1, x_2, x_3$

    - All variables are used in $y$.

- **Noise**:

    - No noise is added ($\epsilon = 0$).

- **Details**:

    - This is a multivariate polynomial function involving cubic and quadratic terms.

    - The function tests the model's ability to capture higher-order polynomial relationships and interactions between variables.

- **Operators Used in Symbolic Regression**:

  - Multiplication ($\times$)

  - Power functions (square, $x^2$; cube, $x^3$)

- **Termination Criteria (Metrics) for SyMANTIC**:

  - RMSE threshold: 0.001

  - $R^2$ threshold: 1.0

## S1.8  Case 8

- **Mathematical Function**:

$$y = x_1^3 + x_1^2 + x_1$$

- **Input Variables**:

  - $x_1, x_2$

  - Only $x_1$ is used in $y$; $x_2$ is generated but not used.

- **Noise**:

  - No noise is added ($\epsilon = 0$).

- **Details**:

  - Known as the Nguyen-1 benchmark function in symbolic regression literature.

  - This univariate polynomial provides a straightforward test case for model accuracy on simpler functional relationships.

- **Operators Used in Symbolic Regression**:

  - Power functions (square, $x^2$; cube, $x^3$)

- Addition (+)

- **Termination Criteria (Metrics) for SyMANTIC**:

  - RMSE threshold: 0.001

  - $R^2$ threshold: 0.99

## S1.9 Case 9

- **Mathematical Function**:

$$y = x_1^4 - x_1^3 + 0.5x_2^2 - x_2$$

- **Input Variables**:

  - $x_1, x_2$

  - Both variables are used in $y$.

- **Noise**:

  - No noise is added ($\epsilon = 0$).

- **Details**:

  - Known as the Nguyen-12 benchmark function in symbolic regression.

  - The function incorporates higher-degree polynomials, presenting a challenge for models to accurately capture the relationship.

- **Operators Used in Symbolic Regression**:

  - Power functions (square, $x^2$; cube, $x^3$; quadrtic, $x^4$)

  - Addition (+)

- **Termination Criteria (Metrics) for SyMANTIC**:
  - RMSE threshold: 0.001
  - $R^2$ threshold: 1.0

## S1.10  Case 10

- **Mathematical Function**:

$$y = \sin(x_1^2)\cos(x_1) - 2$$

- **Input Variables**:
  - $x_1, x_2, x_3, x_4$
  - Only $x_1$ is used in $y$; $x_2, x_3, x_4$ are generated but not used.

- **Noise**:
  - No noise is added ($\epsilon = 0$).

- **Details**:
  - Known as the Livermore-2 benchmark function in symbolic regression.
  - The function involves a product of sine and cosine functions, adding complexity through trigonometric interactions.

- **Operators Used in Symbolic Regression**:
  - Sine function (sin)
  - Cosine function (cos)
  - Power function (square, $x^2$)
  - Multiplication ($\times$)

- **Termination Criteria (Metrics) for SyMANTIC**:
    - RMSE threshold: 0.001
    - $R^2$ threshold: 1.0

**Additional Notes**

- **Unused Variables**: In several cases, additional variables are generated but not used in the target function. This simulates scenarios where datasets contain irrelevant features, testing the model's feature selection and noise robustness capabilities.

- **Noise Levels**:
    - **With Noise**: Cases 1 to 4 include Gaussian noise in the target variable to simulate measurement errors.
    - **Without Noise**: Cases 5 to 10 do not include noise, allowing for the evaluation of model performance under ideal conditions.

The synthetic datasets generated for this study offer a comprehensive set of challenges for evaluating model performance. By encompassing a variety of mathematical functions—including rational expressions, trigonometric functions, exponential decay, and high-degree polynomials—we aim to assess the models' abilities to capture complex relationships, handle irrelevant features, and perform accurately in both noisy and noise-free environments.

These datasets are valuable for benchmarking and can be utilized by other researchers to compare model performance under controlled conditions.

# S2 Details on Science Benchmark Problems

We extend our evaluation by incorporating empirical physical laws from [1] into our synthetic datasets. Each dataset represents a different case with a unique underlying physical formula.

Below, we provide detailed information on how the data for each case is generated, including the mathematical functions used, input variable ranges, noise levels, operators used in symbolic regression, termination criteria, and any additional pertinent details.

## S2.1 Case 11: Hubble's Law

- **Mathematical Function**:

$$v = H_0 D$$

- **Input Variables**:

    - $D$: Distance to a galaxy, linearly spaced between 1 and 20 megaparsecs (Mpc).
    - $H_0$: Hubble constant, set to 73.3 km/s/Mpc.

- **Target Variable**:

    - $v$: Recessional velocity of the galaxy.

- **Noise**:

    - No noise is added ($\epsilon = 0$).

- **Operators Used in Symbolic Regression**:

    - Multiplication ($\times$)

- **Termination Criteria (Metrics)**:

    - RMSE threshold: $1 \times 10^{-8}$
    - $R^2$ threshold: 1.0

## S2.2  Case 12: Leavitt's Law (Period-Luminosity Relation)

- **Mathematical Function**:
$$M = \alpha \log_{10}(P) + \delta$$

- **Input Variables**:
  - $P$: Period of Cepheid variable stars, logarithmically spaced between 1.25 and 158 days.
  - $\alpha$: Slope parameter, set to $-2.084$.
  - $\delta$: Intercept parameter, set to 15.65.

- **Target Variable**:
  - $M$: Absolute magnitude of the star.

- **Noise**:
  - No noise is added ($\epsilon = 0$).

- **Operators Used in Symbolic Regression**:
  - Multiplication ($\times$)
  - Logarithm (base 10) function ($\log_{10}$)

- **Termination Criteria (Metrics)**:
  - RMSE threshold: $1 \times 10^{-8}$
  - $R^2$ threshold: 1.0

- **Details**:
  - Leavitt's Law relates the luminosity of Cepheid variable stars to their pulsation period, critical for distance measurements in astronomy.

- Periods are generated on a logarithmic scale to reflect the wide range of observed periods.

## S2.3   Case 13: Rydberg Formula

- **Mathematical Function**:
$$\frac{1}{\lambda} = R_H \left( \frac{1}{n_1^2} - \frac{1}{n_2^2} \right)$$

- **Input Variables**:
  - $n_1$: Lower energy level, randomly selected integer between 1 and 6.
  - $n_2$: Higher energy level, randomly selected integer greater than $n_1$ and up to 7.
  - $R_H$: Rydberg constant for hydrogen, $1.097 \times 10^7$ m$^{-1}$.

- **Target Variable**:
  - $\frac{1}{\lambda}$: Inverse wavelength of emitted or absorbed light.

- **Noise**:
  - No noise is added ($\epsilon = 0$).

- **Operators Used in Symbolic Regression**:
  - Power function (square, $x^2$)
  - Inverse function ($x^{-1}$)
  - Subtraction ($-$)
  - Division ($/$)
  - Addition ($+$)

- **Termination Criteria (Metrics)**:
  - RMSE threshold: $1 \times 10^{-8}$

– $R^2$ threshold: 1.0

- **Details**:

    – The Rydberg formula predicts the wavelengths of spectral lines of hydrogen.

    – The energy levels $n_1$ and $n_2$ are chosen to ensure valid transitions.

## S2.4  Case 14: Ideal Gas Law

- **Mathematical Function**:
$$P = \frac{nRT}{V}$$

- **Input Variables**:

    – $n$: Amount of substance in moles, randomly sampled between 1 and 100 mol.

    – $T$: Temperature, randomly sampled between 300 and 400 K.

    – $V$: Volume, randomly sampled between 1 and 100 L.

    – $R$: Ideal gas constant, 8.314 J/(mol·K).

- **Target Variable**:

    – $P$: Pressure of the gas.

- **Noise**:

    – No noise is added ($\epsilon = 0$).

- **Operators Used in Symbolic Regression**:

    – Multiplication ($\times$)

    – Division (/)

- **Termination Criteria (Metrics)**:

- RMSE threshold: $1 \times 10^{-8}$

- $R^2$ threshold: 1.0

- **Details**:

  - The Ideal Gas Law relates pressure, volume, temperature, and amount of gas.

## S2.5 Case 15: Newton's Law of Universal Gravitation

- **Mathematical Function**:

$$F = G \frac{m_1 m_2}{r^2}$$

- **Input Variables**:

  - $m_1$: Mass of the first object, randomly sampled between $1.3 \times 10^{22}$ kg and $2 \times 10^{30}$ kg.

  - $m_2$: Mass of the second object, same range as $m_1$.

  - $r$: Distance between the centers of the two masses, randomly sampled between $3.7 \times 10^8$ m and $7.37593 \times 10^{12}$ m.

  - $G$: Gravitational constant, $6.67408 \times 10^{-11}$ N·m$^2$/kg$^2$.

- **Target Variable**:

  - $F$: Gravitational force between the two masses.

- **Noise**:

  - No noise is added ($\epsilon = 0$).

- **Operators Used in Symbolic Regression**:

  - Multiplication ($\times$)

  - Division (/)

- Power function (square, $x^2$)

- **Termination Criteria (Metrics)**:

    - RMSE threshold: $1 \times 10^{-8}$

    - $R^2$ threshold: $1.0$

- **Details**:

    - Newton's Law quantifies the gravitational force between two masses.

    - Masses and distances are chosen to represent a range from planetary to stellar scales.

By including datasets based on fundamental empirical physical laws, we aim to demonstrate the capability of our symbolic regression models to rediscover well-established relationships from data. The absence of noise and the use of appropriate operators allow the models to recover the exact mathematical forms of the laws. These cases provide a rigorous test of the model's effectiveness and its potential for discovering underlying physical principles from observational data.

# S3 Details on Feynman Equations

In addition to the previously discussed synthetic and empirical datasets, we further evaluate our symbolic regression model on a set of equations derived from fundamental physical laws, specifically focusing on complex relationships as presented in the Feynman Lectures on Physics. The datasets are obtained from publicly available sources and are used to test the model's ability to rediscover fundamental physical equations from data.[2]

**Source of Data**

The datasets are sourced from the following repositories:

- SRSD-Feynman Medium Dataset on Hugging Face.

- SRSD Benchmark on GitHub.

We sampled 50 training data points from each dataset to keep it consistent with the low data-regime.

### S3.1 Feynman Equations and Case Studies

We consider the following equations:

1. **Equation A: Distance Formula**

$$d^2 = (x_2 - x_1)^2 + (x_3 - x_4)^2$$

2. **Equation B: Relativistic Mass**

$$m^2 = \frac{m_0^2}{1 - \frac{v^2}{c^2}}$$

3. **Equation C: Position in an Electromagnetic Field**

$$x = \frac{qe}{m(\omega_1^2 - \omega_2^2)}$$

4. **Equation D: Force in an Electromagnetic Field**

$$F = q\left(E + Bv\sin(\theta)\right)$$

5. **Equation E: Gravitational Potential Energy Difference**

$$F = Gm_1m_2\left(\frac{1}{r_2} - \frac{1}{r_1}\right)$$

For each equation, we perform symbolic regression using our model with appropriate operators, dimensional considerations, and termination criteria.

## S3.2 Case 16: Distance Formula

- **Mathematical Function**:

$$d^2 = (x_2 - x_1)^2 + (x_3 - x_4)^2$$

- **Input Variables**:

  - $x_1, x_2, x_3, x_4$: Coordinates in a two-dimensional space.

- **Target Variable**:

  - $d^2$: Square of the distance between two points.

- **Operators Used in Symbolic Regression**:

  - Addition (+)
  - Subtraction (−)
  - Power function (square, $x^2$)
  - Square root function ($\sqrt{\ }$)

- **Dimensionality Constraints**:

  - All variables have the same dimensionality (e.g., length units).

- **Termination Criteria**:

  - Default criteria used.

## S3.3 Case 17: Relativistic Mass

- **Mathematical Function**:

$$m^2 = \frac{m_0^2}{1 - \frac{v^2}{c^2}}$$

- **Input Variables**:

    - $m_0$: Rest mass of the object.

    - $v$: Velocity of the object.

    - $c$: Speed of light.

- **Target Variable**:

    - $m^2$: Square of the relativistic mass.

- **Operators Used in Symbolic Regression**:

    - Division (/)

    - Multiplication (×)

    - Power function (square, $x^2$)

    - Subtraction (−)

- **Dimensionality Constraints**:

    - Variables have dimensions of mass and velocity.

- **Termination Criteria**:

    - RMSE threshold: 0.001

    - $R^2$ threshold: 1.0

## S3.4 Case 18: Position in an Electromagnetic Field

- **Mathematical Function**:

$$x = \frac{qe}{m(\omega_1^2 - \omega_2^2)}$$

- **Input Variables**:

  - $q$: Charge of the particle.

  - $e$: Elementary charge.

  - $m$: Mass of the particle.

  - $\omega_1, \omega_2$: Angular frequencies.

- **Target Variable**:

  - $x$: Position of the particle.

- **Operators Used in Symbolic Regression**:

  - Division (/)

  - Multiplication (×)

  - Power function (square, $x^2$)

  - Subtraction (−)

- **Dimensionality Constraints**:

  - Units of charge, mass, and angular frequency are considered to guide the regression.

- **Termination Criteria**:

  - Default criteria used.

## S3.5 Case 19: Force in an Electromagnetic Field

- **Mathematical Function**:

$$F = q\left(E + Bv\sin(\theta)\right)$$

- **Input Variables**:

  - $q$: Charge of the particle.
  - $E$: Electric field strength.
  - $B$: Magnetic field strength.
  - $v$: Velocity of the particle.
  - $\theta$: Angle between the velocity and the magnetic field.

- **Target Variable**:

  - $F$: Force experienced by the particle.

- **Operators Used in Symbolic Regression**:

  - Addition (+)
  - Multiplication (×)
  - Sine function (sin)

- **Dimensionality Constraints**:

  - Units of charge, electric field, magnetic field, velocity, and angle are considered.

- **Termination Criteria**:

  - RMSE threshold: 0.001
  - $R^2$ threshold: 1.0

## S3.6 Case 20: Gravitational Potential Energy Difference

- **Mathematical Function**:

$$F = Gm_1m_2 \left( \frac{1}{r_2} - \frac{1}{r_1} \right)$$

- **Input Variables**:
  - $G$: Gravitational constant.
  - $m_1, m_2$: Masses of the two objects.
  - $r_1, r_2$: Distances from the masses to a reference point.

- **Target Variable**:
  - $F$: Gravitational potential energy difference.

- **Operators Used in Symbolic Regression**:
  - Multiplication ($\times$)
  - Inverse function ($x^{-1}$)
  - Subtraction ($-$)

- **Dimensionality Constraints**:
  - Variables have dimensions of mass and length.

- **Termination Criteria**:
  - RMSE threshold: $1 \times 10^{-8}$
  - $R^2$ threshold: 1.0

By testing our symbolic regression model on these complex physical equations from the Feynman Lectures, we aim to evaluate its capability to rediscover fundamental physical laws

from data. The inclusion of dimensional analysis, appropriate mathematical operators, and strict termination criteria enhances the model's ability to generate physically meaningful and accurate equations. These case studies demonstrate the potential of symbolic regression in uncovering underlying physical principles from experimental or simulated data.

## S4 Impact of Noise on Lorentz System

The Lorenz system is a canonical example in the study of chaotic dynamics and nonlinear systems. In this section, we compare the performance of three methods—PySINDy, Ensemble SINDy, and our proposed method (SyMANTIC) — for identifying governing equations from simulated data. The aim is to evaluate accuracy (reconstruction of structure) and robustness to noise.

### Data Generation:

The training data for the Lorenz system was generated by numerically solving the following equations:

$$\frac{dx}{dt} = \sigma(y - x), \tag{1}$$

$$\frac{dy}{dt} = x(\rho - z) - y, \tag{2}$$

$$\frac{dz}{dt} = xy - \beta z, \tag{3}$$

where the parameters were set as $\sigma = 10$, $\rho = 28$, and $\beta = \frac{8}{3}$. The system was integrated using the LSODA solver with absolute and relative tolerances of $10^{-12}$. The integration was performed over the time span $t \in [0, 100]$ with a timestep $\Delta t = 0.001$. The initial conditions were $x(0) = -8$, $y(0) = 8$, and $z(0) = 27$.

To simulate experimental noise, Gaussian noise with a standard deviation ($\sigma = 0.01, 0.1, 1$)

was added to the true time derivatives of each variable:

$$\dot{x} = \dot{x}_{\text{true}} + \eta_x, \quad \eta_x \sim \mathcal{N}(0, \sigma),$$

$$\dot{y} = \dot{y}_{\text{true}} + \eta_y, \quad \eta_y \sim \mathcal{N}(0, \sigma),$$

$$\dot{z} = \dot{z}_{\text{true}} + \eta_z, \quad \eta_z \sim \mathcal{N}(0, \sigma).$$

The test data was generated using the same Lorenz equations, parameters, and numerical solver as the training data. However, the initial conditions for testing were set as $x(0) = 8$, $y(0) = 7$, and $z(0) = 15$, and the integration was performed over the time span $t \in [0, 15]$ with $\Delta t = 0.001$.

## Hyperparameters:

To ensure a fair comparison, we restricted our analysis to interaction parameters for PySINDy and E-SINDy, while focusing exclusively on the $[*]$ operator for SyMANTIC for the feature space. All other hyperparameters across these methods were maintained at their respective default settings, ensuring consistency in benchmarking and evaluation.

## Results:

The results illustrate the consistent superiority of the SyMANTIC framework at varying noise levels, as shown in Figure S1. In scenarios with significant noise ($\sigma = 1.0$), SyMANTIC demonstrates robustness and precision in capturing underlying patterns, outperforming PySINDy and E-SINDy.

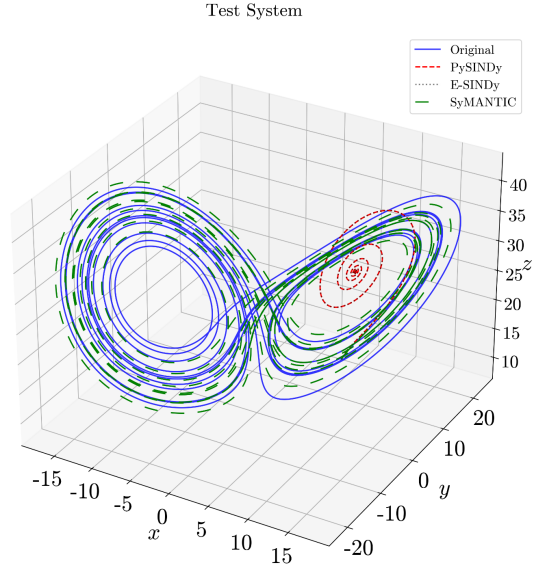

(a) Low noise ($\sigma = 0.01$)

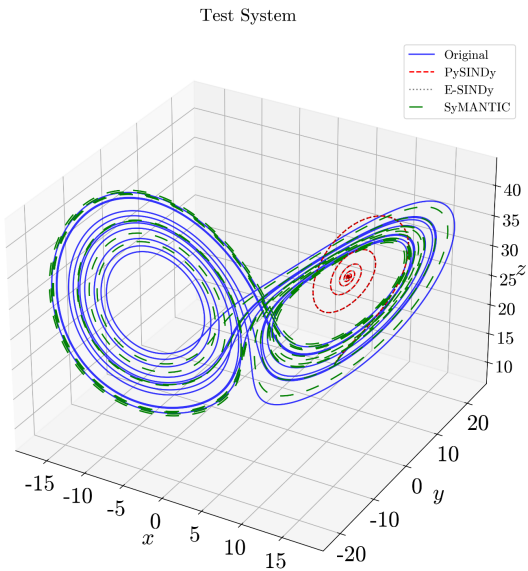

(b) Moderate noise ($\sigma = 0.1$)

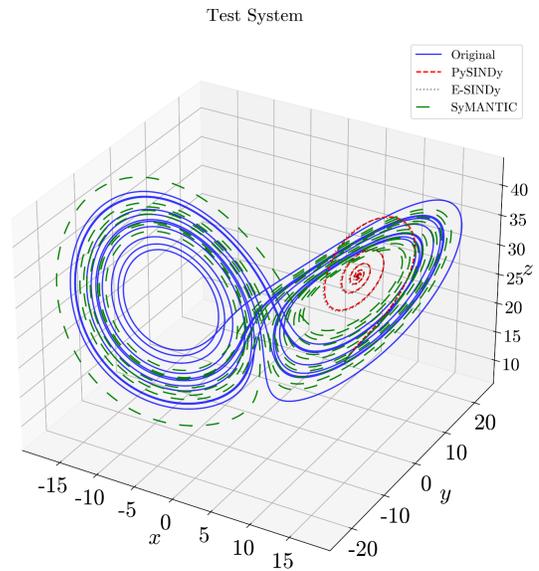

(c) High noise ($\sigma = 1.0$)

Figure S1. Comparison of SyMANTIC performance across varying noise levels.

In our extended case study with a larger dataset (100 time points), we observed that as the number of data points increased, PySINDy approached performance levels comparable to SyMANTIC. Despite its initial limitations with smaller datasets or noisy data, PySINDy demonstrated a convergence toward the performance of SyMANTIC as more data became

27

available.

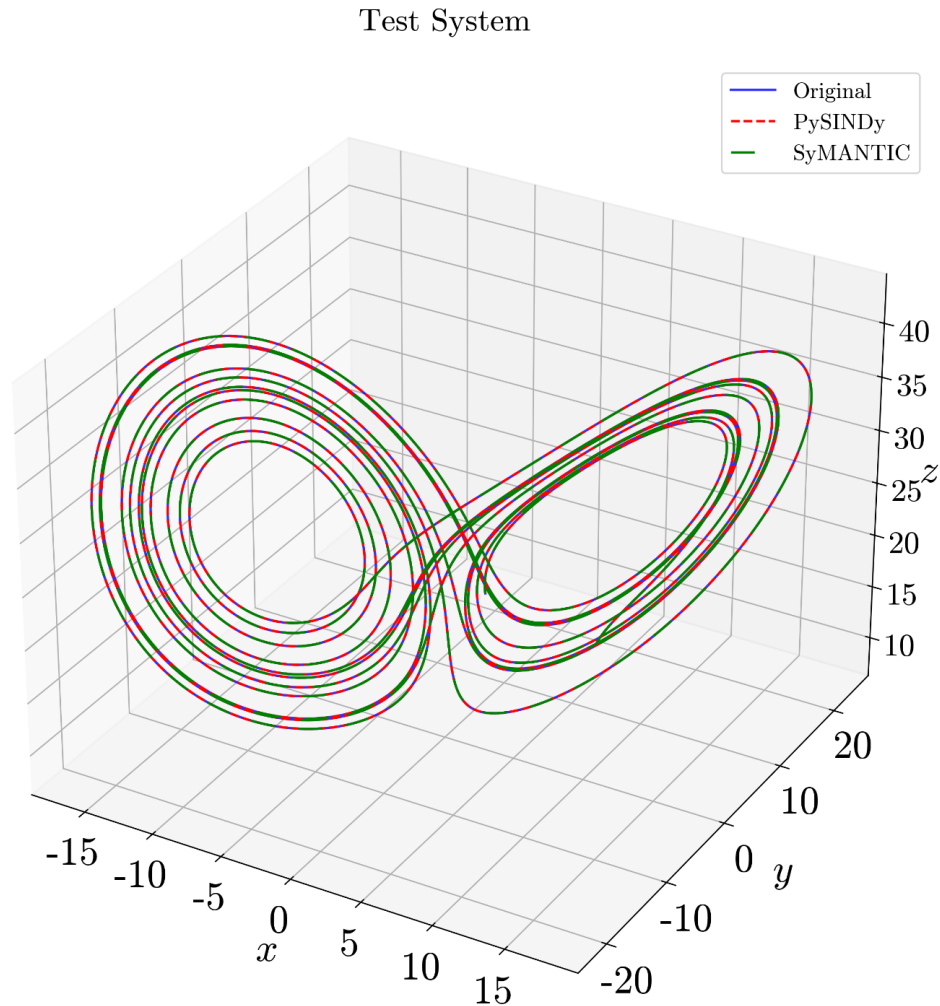

Figure S2: Comparison between SyMANTIC and PySINDy using data collected from 100 timepoints.

# S5   Comparing SyMANTIC to Classical Feature Selection Methods

In this case study, we compared SyMANTIC (Model building Screening + $L_0$) performance with classical feature selection techniques such as Forward Feature Selection (FFS) and Backward Feature Selection (BFS) in building a four-term model:

$$y = (x_1 + x_2 + x_3 + x_4) + N(0, 1)$$

In Forward Feature Selection (FFS), the model starts with no features and adds them one by one, selecting the feature that improves the model's performance the most at each step. In Backward Feature Selection (BFS), the model starts with all features and iteratively removes the least significant features.

Although FFS and BFS are effective in classical regression contexts, they can become computationally expensive, especially when the dataset contains a large number of features. This is clearly evident from the results in the table below. In comparison, SyMANTIC (model building) demonstrates superior performance, with significantly lower computational cost, even as the number of features increases.

Table S1: Comparison of computational time (in seconds) for Forward Feature Selection (FFS), Backward Feature Selection (BFS), and SyMANTIC in building a 4-term model with increasing feature sizes.

| Number of Features | FFS | BFS | SyMANTIC Approach |
|---|---|---|---|
| 1000 | 13.17 | 97.32 | 0.37 |
| 10000 | 39.11 | 432.31 | 1.24 |
| 100000 | 197.56 | 3298.39 | 1.98 |

Note: This limitation becomes particularly pronounced when these methods are employed recursively within the C$^2$-SISSO framework.

# TOC Graphic

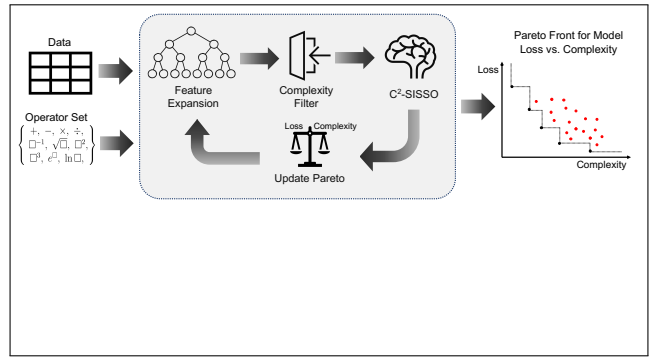



\end{document}